\documentclass[10pt,journal,compsoc]{IEEEtran}

\usepackage{times}
\usepackage{epsfig}
\usepackage{graphicx}
\usepackage{subfigure}
\usepackage{amsmath}
\usepackage{amssymb}
\usepackage{mathrsfs}
\usepackage{bm}
\usepackage{overpic}
\usepackage{multirow}
\usepackage{booktabs}       
\usepackage{array}
\usepackage{paralist}
\usepackage{makecell}
\newcommand{\PreserveBackslash}[1]{\let\temp=\\#1\let\\=\temp}
\newcolumntype{C}[1]{>{\PreserveBackslash\centering}p{#1}}
\newcolumntype{R}[1]{>{\PreserveBackslash\raggedleft}p{#1}}
\newcolumntype{L}[1]{>{\PreserveBackslash\raggedright}p{#1}}
\newcommand{\Paragraph}[1]{\noindent\textbf{#1}}

\usepackage[colorlinks,
linkcolor=blue,
anchorcolor=blue,
citecolor=blue,
]{hyperref}
\usepackage{url}

\ifCLASSOPTIONcompsoc
\usepackage[nocompress]{cite}
\else
\usepackage{cite}
\fi

\def\etal{\emph{et al}.~}
\def\ie{i.e.,~} 
\def\eg{\emph{e.g}}

\def\blu#1{\textbf{\color{blue} #1}} 
\def\red#1{\textbf{\color{red}\underline{#1}}} 

\ifCLASSINFOpdf
\else
\fi

\begin{document}
	%
	\title{Learning Selective Mutual Attention and Contrast for RGB-D Saliency Detection}

	\author{Nian~Liu,~\IEEEmembership{Member,~IEEE,}
		Ni~Zhang,
		Ling~Shao,~\IEEEmembership{Senior~Member,~IEEE,}
		and~Junwei~Han,~\IEEEmembership{Senior~Member,~IEEE}
		\IEEEcompsocitemizethanks{\IEEEcompsocthanksitem N. Liu is with the Mohamed bin Zayed University of Artificial Intelligence, Abu Dhabi, UAE. E-mail: liunian228@gmail.com
		\IEEEcompsocthanksitem N. Zhang and J. Han are with School of Automation, Northwestern Polytechnical University, Xi'an, China, E-mail: \{nnizhang.1995,junweihan2010\}@gmail.com
		\IEEEcompsocthanksitem L. Shao is with the Mohamed bin Zayed University of Artificial Intelligence, Abu Dhabi, UAE, and also with the Inception Institute of Artificial Intelligence, Abu Dhabi, UAE. E-mail: ling.shao@inceptioniai.org
		\IEEEcompsocthanksitem N. Liu and N. Zhang contribute equally to this paper.
	}
	}

	\markboth{IEEE Transactions on Pattern Analysis and Machine Intelligence}%
	{Shell \MakeLowercase{\textit{et al.}}: Bare Demo of IEEEtran.cls for Computer Society Journals}

	\IEEEtitleabstractindextext{%
		\begin{abstract}
			How to effectively fuse cross-modal information is the key problem for RGB-D salient object detection. Early fusion and the result fusion schemes fuse RGB and depth information at the input and output stages, respectively, hence incur the problem of distribution gap or information loss. Many models use the feature fusion strategy but are limited by the low-order point-to-point fusion methods. In this paper, we propose a novel mutual attention model by fusing attention and contexts from different modalities. We use the non-local attention of one modality to propagate long-range contextual dependencies for the other modality, thus leveraging complementary attention cues to perform high-order and trilinear cross-modal interaction. We also propose to induce contrast inference from the mutual attention and obtain a unified model. Considering low-quality depth data may detriment the model performance, we further propose selective attention to reweight the added depth cues. We embed the proposed modules in a two-stream CNN for RGB-D SOD. Experimental results have demonstrated the effectiveness of our proposed model. Moreover, we also construct a new challenging large-scale RGB-D SOD dataset with high-quality, thus can both promote the training and evaluation of deep models.
		\end{abstract}
		
		\begin{IEEEkeywords}
			salient object detection, RGB-D image, attention model, contrast, non-local network.
	\end{IEEEkeywords}}

	\maketitle

	\IEEEraisesectionheading{\section{Introduction}\label{sec:introduction}}
	
	
	\IEEEPARstart{O}{ver} the past few decades, researchers have proposed many computational salient object detection (SOD) models and achieved very promising performance, \eg, \cite{liu2016dhsnet,liu2020picanetTIP,wang2019iterative,zhao2019egnet,qin2019basnet}. However, most of them work on RGB images and only leverage appearance cues, which usually incur insurmountable difficulties in many challenging scenarios.
	In the meantime, as human beings, we live in a real 3D environment. Our visual system heavily relies on depth information, which can supply sufficient complementary cues for the appearance information. Thus, it is quite natural and necessary to incorporate both RGB and depth data for solving the SOD problem.
	
	To combine these two modalities for RGB-D SOD, some existing works employ the \textbf{early fusion} strategy \cite{song2017mdsf,qu2017df,liu2019se,fan2019d3net}, with which SOD models take both RGB and depth data as inputs and process them in a unified way. However, this kind of model faces the difficulty of using one model to fit the data from two modalities well due to their distribution gap. Some other models \cite{desingh2013depth,guo2016salient,wang2019afnet} adopt two submodules to generate saliency maps for the RGB data and the depth data separately, and then use a fusion method to combine the two saliency maps. This kind of strategy is called \textbf{result fusion}. These methods are also suboptimal since using two separate SOD modeling processes will gradually compress and lose rich modality information. As a result, cross-modal interaction between the two saliency maps is highly limited.
	
	Many other RGB-D SOD works exploit the \textbf{middle fusion} strategy as a better choice, which first fuses intermediate information of the two modalities and then generates a final saliency map. Most typically, many recent models \cite{Piao2019dmra,chen2019mmci,shigematsu2017learning,han2017ctmf,chen2018pcf,fu2020jl,zhang2020ssf,chen2019tanet} first extract RGB and depth features separately using two-stream CNNs, and then combine the cross-modal features in decoders. 
	We generalize these methods as belonging to the \textbf{feature fusion} strategy.
	Although it avoids the distribution discrepancy problem and fuses rich multi-modal features, the feature fusion methods are usually simple summation, multiplication, or concatenation, without exploring more powerful multi-modal feature interaction.
	
\begin{figure}[t]
		\graphicspath{{Figures/Figure1/}}
		\centering
		\includegraphics[width=1\linewidth]{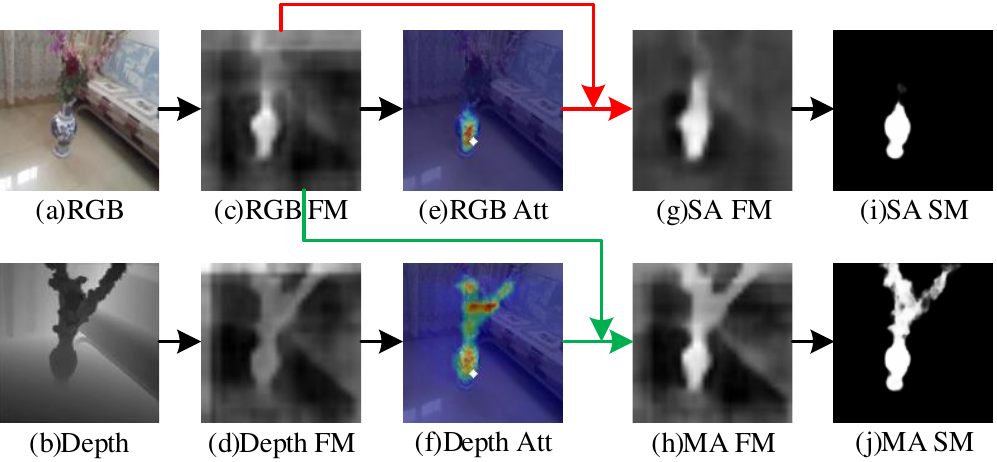}
		\caption{\textbf{Comparison on the effectiveness of using self-attention (SA) and mutual attention (MA).} We first give the RGB image and the depth map of an example image pair in (a) and (b). Then, we show the feature maps (FM) of the two modalities in (c) and (d). In (e) and (f), we show the attention maps (Att) of a query position (the white point) for the two modalities. Next, we adopt the self-attention (SA) mechanism (shown as {\color{red} red} paths), which uses ``RGB Att" to propagate context features on ``RGB FM", obtaining the feature map ``SA FM" and the final saliency map ``SA SM". As a contrast, we also adopt the proposed mutual-attention (MA) mechanism (shown as {\color{green} green} paths), which uses ``Depth Att" to propagate context features on ``RGB FM", obtaining ``MA FM" and ``MA SM". We observe that the mutual-attention mechanism can offer a different guidance for context propagation and obtain better SOD results.}
		\label{figure1}
		\vspace{-0.4cm}
	\end{figure}

	In this paper, we propose a novel and more effective middle fusion strategy. Inspired by the Non-local (NL) network \cite{wang2018nonlocal}, we propose to exploit cross-modal attention propagation. The NL network first generates long-range spatial attention for each query position via computing the query-key pair-wise affinity,
	and then uses the attention to propagate global context features. Given that the attention and the propagated features are both induced from the same feature map, the NL model is regarded as belonging to the \textbf{self-attention} mechanism. However, since the attention mainly activates on regions similar to the query position, the NL model is very sensitive to the feature quality. If the input feature map is not discriminative enough, the propagated context features from similar regions can only supply limited information gain for each query position.
	An intuitive example is given in Figure~\ref{figure1}. We can see from (c) that the RGB feature map only has large activations on the vase and ignores most parts of the flower. Hence, for a query position located on the body of the vase, which is marked as the white point in (e), its self-attention map mainly activates on the vase. As a result, the feature map after using self-attention shown in (g) and the resultant saliency map shown in (i) still miss most parts of the flower.
	
	Considering the property that RGB and depth data can complement each other, we propose to propagate global context using each other's attention, to which we refer as the \textbf{mutual-attention} mechanism. It supplies complementary cues about where should attend based on the information of the other modality. When the depth attention is used for context propagation in the RGB modality, the attended regions are not limited to those that have similar appearance with the query position anymore. Instead, they correspond to those who have similar depth cues, thus providing additional informative contexts for the query position. The same goes for the RGB attention. The example in Figure~\ref{figure1}(f) shows that for the given query point, the depth attention can learn to attend on the whole foreground object instead of only highlighting the vase. Finally, the flower region can be enhanced in the feature map and detected in the saliency map. We also show that the proposed mutual attention mechanism actually introduces high-order and trilinear information interactions for the RGB and depth modality. Compared with previous fusion methods, our model thus has more powerful multi-modal learning capability.
	
	Furthermore, since SOD aims to find distinct objects in each image, it naturally involves contrast inference, which has been widely used in previous saliency models, \eg \cite{itti1998model,cheng2014gc,li2016deep}. Therefore, we integrate this mechanism by inferring the contrast regions from the computed mutual attention. As a result, contrast inference can be unified with the attention model without many extra computational costs. We adopt the novel mutual attention and contrast model in a two-stream U-shape \cite{ronneberger2015unet} network to fuse multi-modal cues at multiple scales for RGB-D SOD. Since the depth information serves as complementarity for the RGB feature and many depth maps are of low-quality, we also propose selective attention to decide how much depth cues should be involved in our model. Experimental results successfully verify the effectiveness of our proposed model.
	
	Another urgent issue in RGB-D SOD is the lack of high-quality and large-scale benchmark datasets. Although there are as many as eight datasets (\ie \cite{niu2012stere,li2014lfsd,cheng2014rgbd135,peng2014nlpr,ju2014njud,zhu2017ssd,Piao2019dmra,fan2019d3net}) widely used in previous works for benchmarking, most of them have simplex scenes and objects, or insufficient images, or low-quality depth maps. These issues not only limit the training of models with good performance and generalization ability, but also hinder a comprehensive performance evaluation. To this end, we construct a large-scale RGB-D SOD dataset with diverse real-life scenes and good depth quality. We have released the dataset\footnote[1]{\url{https://github.com/nnizhang/SMAC}} and believe it can benefit the RGB-D SOD research community much.
	
	To sum up, the contributions of our work can be summarized as follows:
	\begin{compactitem}
		\item We present a novel mutual attention model for multi-modal information fusion in RGB-D SOD. It leverages complementary attention knowledge in long-range context propagation and introduces high-order and trilinear modality interactions. We also propose to unify the contrast mechanism in it.
		\item We adopt the proposed model in a two-stream UNet for RGB-D SOD. In the decoders, mutual attention can be further used for multi-scale cross-modal interaction. We also propose selective attention to weight the fused depth cues thus reduce the distraction of low-quality depth information.
		\item Aiming at the problem of lack of high-quality and large-scale benchmark datasets, we construct a new RGB-D SOD dataset with the most image-pairs, the most diverse visual scenes, and high-quality depth maps. It can help train deep models with better generalization and achieve a more comprehensive evaluation.
		\item We conduct performance evaluation on nine RGB-D SOD benchmark datasets. Experimental results verify the effectiveness of our proposed models, especially the mutual attention model. Finally, our overall RGB-D SOD model performs favorably against previous state-of-the-art methods. \\
	\end{compactitem}
	
	Compared with our previous version of this work \cite{liu2020s2ma}, we have made the following extensions. First, we found self-attention was not beneficial when fused with mutual attention as in \cite{liu2020s2ma}, which we argue is due to the low-quality input feature maps. Thus we do not fuse self-attention and mutual attention anymore. Instead, we fuse mutual attention and contrast. Then we find it useful to cascade a self-attention model right after using mutual attention and contrast. We believe it is because the feature maps have been largely promoted and become more discriminative after using mutual attention and contrast. Second, in \cite{liu2020s2ma} we only used the $S^2$MA module right after the encoders. In this work, we find that further using mutual attention in subsequent decoders can improve the model performance. Third, we propose a new and more effective selective attention method in this work and adopt it in every cross-modal fusion module. Forth, we construct a new large-scale, and high-quality RGB-D SOD benchmark dataset.

	\section{Related Work}
	\subsection{Saliency Detection on RGB-D Images}
	Traditional RGB-D SOD methods usually borrow common priors (\eg, contrast \cite{cheng2014rgbd135} and compactness \cite{Cong2016Saliency}) from RGB SOD models to design RGB and depth features. Additionally, some researchers proposed to exploit depth-specific priors,
	\eg, shape and 3D layout priors \cite{Ciptadi2013An}, and anisotropic center-surround difference \cite{ju2014njud,guo2016salient}.
	
	The aforementioned methods all rely heavily on hand-crafted features and lack high-level representations, which are very important for understanding challenging scenarios. To this end, many recent works introduce CNNs into RGB-D SOD and have achieved promising results. Qu \etal \cite{qu2017df} adopted the early fusion strategy and serialized hand-crafted RGB and depth features together as the CNN inputs. Fan \etal \cite{fan2019d3net} and Liu \etal \cite{liu2019se} used each depth map as the $4^{th}$ channel along with the corresponding RGB image as the CNN input. In \cite{wang2019afnet}, Wang \etal adopted the result fusion strategy and adaptively fused RGB and depth saliency maps with a learned switch map. Recently, the middle fusion strategy is adopted by many works to fuse intermediate depth and appearance features. Han \etal \cite{han2017ctmf} fused the fully-connected representations of the RGB and the depth branches into a joint representation. Most other models \cite{chen2018pcf,Piao2019dmra,zhang2020ssf,fu2020jl,li2020icnet} fused complementary cross-modal convolutional features at multiple scales by various methods, such as summation, multiplication, and concatenation. In contrast, our model fuses cross-modal non-local attention and context features, thus introducing high-order and trilinear information interactions.
	
	The attention mechanism is also widely used in existing works to fuse RGB and depth modalities, \eg, in \cite{zhao2019cpfp} and \cite{Piao2019dmra}. However, they only generated channel \cite{Piao2019dmra} or spatial \cite{zhao2019cpfp} attention from the depth view and adopted them to filter the appearance features. Nevertheless, we generate non-local attention from both views and then use them to propagate long-range contexts for each other.
	
	
	\subsection{Self-Attention and Multi-modal Attention}
	In \cite{vaswani2017attention}, Vaswani \etal proposed a self-attention network for natural language modeling. Given a query word and a set of key-value pairs for other words, they computed attention weights for all query-key pairs and then aggregated all the values as the context feature. Similarly, Wang \etal \cite{wang2018nonlocal} proposed the NL model for learning self-attention in 2D or 3D vision modeling. For multi-modal attention learning, Nam \etal \cite{nam2017dual} proposed to learn both visual and textual attention mechanisms for multi-modal reasoning and matching. Wan \etal \cite{wan2019multi} applied three attention models in three modalities of source code for the code retrieval task. However, both of them learn and adopt attention for each modality separately, and then fuse the obtained attended features. In \cite{zhang2019pattern} and our previous work \cite{liu2020s2ma}, cross-task and cross-modal attention affinities are fused. Different from them, we propose the mutual attention mechanism to mix up attention and values from different modalities. We also find it more helpful to cascade self-attention right after mutual attention instead of fusing their affinities.
	
	\subsection{RGB-D SOD Datasets} 
	So far, researchers have proposed several RGB-D saliency datasets and eight of them are widely used in RGB-D SOD papers. Niu \etal \cite{niu2012stere} collected and annotated the first stereoscopic SOD dataset \textbf{STERE}. It contains 1,000 pairs of stereoscopic Internet images, most of which have various outdoor scenes and objects. The depth maps are generated using the SIFT flow \cite{liu2010sift} algorithm. Li \etal \cite{li2014lfsd} constructed a light field SOD dataset \textbf{LFSD} using the Lytro light field camera. This dataset includes 100 indoor and outdoor images and the depth maps are directly generated by the Lytro desktop. Most images contain simple foreground objects but have complex or similar backgrounds. The \textbf{NLPR} \cite{peng2014nlpr} dataset and the \textbf{RGBD135} \cite{cheng2014rgbd135} dataset contain 1,000 and 135 images collected by the Microsoft Kinect, respectively. Hence, their depth maps have good quality. However, most of their images include relatively simple foreground objects and visual scenes. The \textbf{NJUD} \cite{ju2014njud} dataset has 1,985 stereo images collected from the Internet, 3D movies, and photos taken by a Fuji W3 camera. Most images show diverse outdoor scenes and foreground objects. The depth maps are generated using Sun’s optical flow method \cite{sun2010secrets}. \textbf{SSD} \cite{zhu2017ssd} is a small-scale dataset which only has 80 stereo movie frames. The images contain several movie scenes with persons, animals, buildings, etc, as foreground objects. The depth maps are also generated using \cite{sun2010secrets}. \textbf{DUTLF-Depth} \cite{Piao2019dmra} is a recently proposed dataset that contains 1,200 images captured by a Lytro2 light field camera. It includes divers complex scenes, e.g., multiple or transparent objects, complex backgrounds, and low-intensity scenes. The authors adopted the method of \cite{tao2013depth} to obtain the depth maps. Fan \etal \cite{fan2019d3net} proposed a Salient Person (\textbf{SIP}) dataset with 929 images to emphasize persons in real-world scenes. All the images and depth maps are collected by a Huawei Mate10 smartphone using its dual camera. A summarization of these datasets is shown in Table~\ref{dataCmpTab}.
	
	\begin{table*} [t]
		\begin{center}
			\caption{\textbf{Statistical comparison of different RGB-D SOD benchmark datasets.} We compare our proposed ReDWeb-S dataset with previous ones in terms of publication year (Year), Image Number (IN), Scene Number (SN), Object Number (ON), depth sensor, Depth Quality (DQ), Rgb Global Contrast (RGC), Depth Global Contrast (DGC), Rgb Interior Contrast (RIC), Depth Interior Contrast (DIC), Center Bias Index (CBI), and Object Size (OS). We use $\uparrow$ to denote a metric is better when the value is larger and $\downarrow$ means the smaller the metric is, the better.}
			\label{dataCmpTab}
			\footnotesize
			\begin{tabular}{@{}R{2.4cm}|R{0.45cm}|R{0.55cm}|R{0.45cm}R{0.45cm}|R{3.9cm}R{0.65cm}|R{0.8cm}R{0.8cm}|R{0.45cm}R{0.45cm}|R{0.75cm}R{0.45cm}}
				\toprule
				\multirow{3}{*}{Datasets} & \multirow{3}{*}{Year} &  & \multicolumn{2}{c|}{Scene Stat} & \multicolumn{2}{c|}{Depth Stat} & \multicolumn{2}{c|}{GC} & \multicolumn{2}{c|}{IC} & & \\ \cmidrule{4-11}
				& & \multicolumn{1}{c|}{IN} & \multicolumn{1}{c}{SN}  & \multicolumn{1}{c|}{ON} & \multicolumn{1}{c}{Depth Sensor} & \multicolumn{1}{c|}{DQ} & \multicolumn{1}{c}{RGC} & \multicolumn{1}{c|}{DGC} & \multicolumn{1}{c}{RIC} & \multicolumn{1}{c|}{DIC} & \multicolumn{1}{c}{CBI} & \multicolumn{1}{c}{OS} \\
				& & \multicolumn{1}{c|}{$\uparrow$} & \multicolumn{1}{c}{$\uparrow$} & \multicolumn{1}{c|}{$\uparrow$} & & \multicolumn{1}{c|}{$\uparrow$} & \multicolumn{1}{c}{$\downarrow$}  & \multicolumn{1}{c|}{$\downarrow$} & \multicolumn{1}{c}{$\uparrow$} & \multicolumn{1}{c|}{$\uparrow$} & \multicolumn{1}{c}{$\uparrow$} &  \\ \midrule
				STERE \cite{niu2012stere} &
				2012  & 1000  & 243   & 310   & Stereo images+sift flow \cite{liu2010sift} & 0.9770  & 287.35 & 407.16 & 4.22  & 1.13  & 103.76 & 0.21 \\
				LFSD \cite{li2014lfsd} &
				2014  & 100   & 52    & 63    & Lytro light field camera \cite{ng2005light} & 0.9826 & 230.10 & 290.36 & 4.37  & 1.01  & 118.30 & 0.27 \\
				RGBD135 \cite{cheng2014rgbd135} &
				2014  & 135   & 48    & 41    & Microsoft Kinect \cite{zhang2012kinect} & 0.9866 & 51.12 & 136.39 & 3.61  & 0.89  & 80.86 & 0.13 \\
				NLPR \cite{peng2014nlpr} &
				2014  & 1000  & 175   & 237   & Microsoft Kinect \cite{zhang2012kinect}  & 0.9896 & 148.78 & 42.59 & 3.54  & 1.01  & 84.16 & 0.13 \\
				NJUD \cite{ju2014njud} &
				2014  & 1985  & 282   & 359   & Stereo image+Sun's \cite{sun2010secrets} & 0.9852 & 152.31 & 254.55 & 4.22  & 1.29  & 114.32 & 0.24 \\
				SSD \cite{zhu2017ssd} &
				2017  & 80    & 38    & 23    & Stereo movies+Sun's \cite{sun2010secrets} & 0.9845 & 494.92 & 796.82 & 3.42  & 1.20  & 139.44 & 0.21 \\
				DUTLF-Depth \cite{Piao2019dmra} &
				2019  & 1200  & 191   & 291   & Lytro2 light field camera+\cite{tao2013depth} & 0.9852 & 18.53 & 44.48 & 3.94  & 1.29  & 123.29 & 0.23 \\
				SIP \cite{fan2019d3net} &
				2019  & 929   & 69    & 1     & Huawei Mate10's dual camera & 0.9920 & 98.03 & 79.90 & 3.05  & 0.73  & 106.25 & 0.20 \\
				\midrule
				\multirow{2}{*}{ReDWeb-S (Ours)} &
				\multirow{2}{*}{-} & \multirow{2}{*}{3179}  & \multirow{2}{*}{332}   & \multirow{2}{*}{432}   & Web stereo images+Flownet2.0 \cite{ilg2017flownet}+post-processing & \multirow{2}{*}{0.9904} & \multirow{2}{*}{82.78} & \multirow{2}{*}{53.10} & \multirow{2}{*}{4.38}  & \multirow{2}{*}{1.60}  & \multirow{2}{*}{148.72} & \multirow{2}{*}{0.27}  \\
				\bottomrule
			\end{tabular}
			\vspace{-0.4cm}
		\end{center}{}
	\end{table*}
	
	Although eight benchmark datasets seem many enough for the RGB-D SOD research, we argue that most of them are unsatisfactory due to three points. First, they do not have diverse enough visual scenes for effectively training and comprehensively evaluating SOD models. Many of them only have simplex salient objects and similar background scenes.  Second, most of them have insufficient images, thus being unsuitable for training modern deep learning-based models. From Table~\ref{dataCmpTab} we can see that only two in the eight datasets have more than 1,000 images. Hence most works use two or three datasets together to train deep networks. Third, since many stereo image-based datasets used early flow-based algorithms to generate depth maps, their depth map quality is usually low due to inaccurate pixel matching, noises in faraway areas, and over-smooth segmentation. On the contrary, the proposed dataset totally includes 3,179 images with various visual scenes and high-quality depth maps, which can promote both training and evaluation of deep RGB-D SOD models.
	
	\section{Proposed Dataset}
	In this paper, we construct a new large-scale challenging RGB-D SOD dataset based on the ReDWeb \cite{xian2018monocular} dataset. It is a state-of-the-art dataset proposed for monocular image depth estimation. It contains 3,600 images selected from many web stereo images with various real-life scenes and foreground objects. For depth map generation, the authors first adopted the state-of-the-art Flownet2.0 algorithm \cite{ilg2017flownet} to generate correspondence maps, and then used a deep semantic segmentation model \cite{lin2017refinenet} to remove noises in sky areas as the post-process, thus resulting in high-quality depth maps. Consequently, ReDWeb supplies a good basis for constructing our high-quality RGB-D SOD dataset, which we name as ReDWeb-S. We elaborate on the dataset construction method and the statistic analysis in this part.
	
	\begin{figure*}[!t]
		\graphicspath{{Figures/scene_object_stat/}}
		\centering
		\includegraphics[width=1\linewidth]{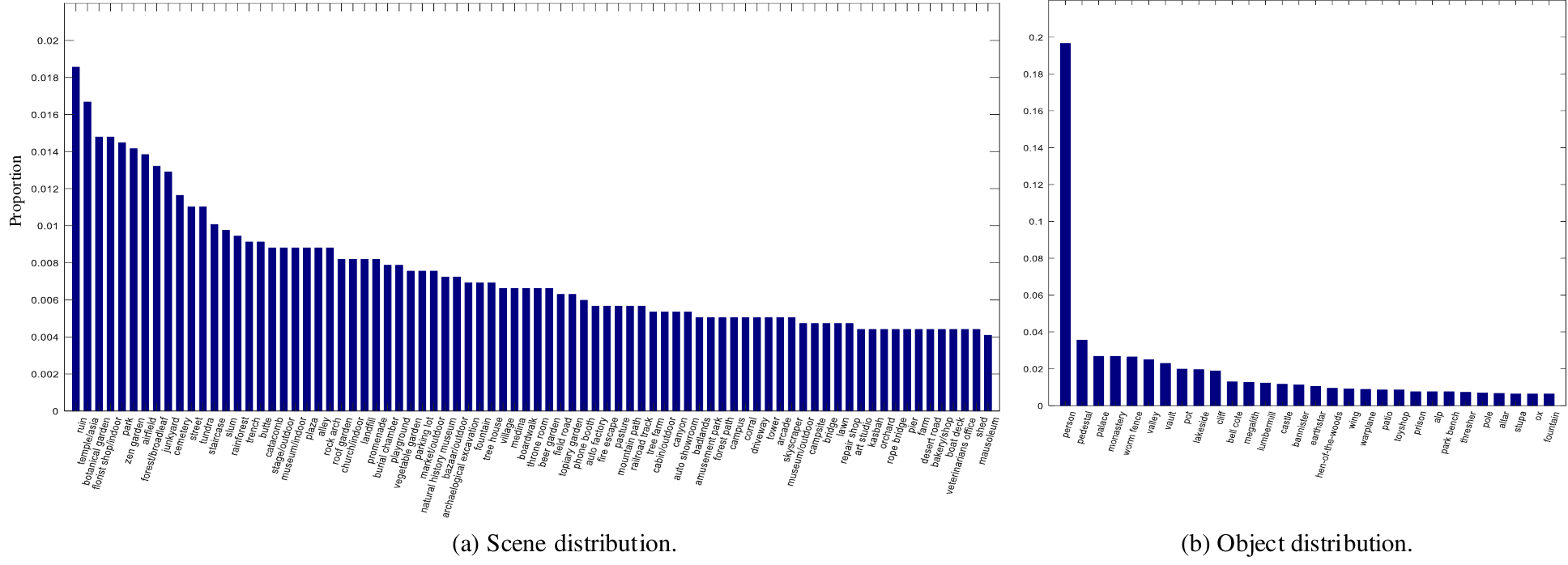}
		\caption{\textbf{Top 60\% scene and object category distributions of our proposed ReDWeb-S dataset.}}
		\label{fig:scene_obj_stat}
		\vspace{-0.3cm}
	\end{figure*}
	
	\subsection{Dataset Construction}
	We first ask four participants to annotate salient objects in each image of the ReDWeb dataset using bounding-boxes (bbox) and also remove images without foreground objects. Then, for each image, we calculate the IoU matching scores for the bboxes of every two annotators and select the bbox with the highest total matching score as the annotation. At last, we ask participants to accurately segment salient objects based on the selected bboxes. As a result, we obtain 3,179 images with both high-quality depth maps and annotated saliency maps. We further randomly split them into a training set with 2,179 RGB-D image pairs and a testing set with the remaining 1,000 image pairs.
	
	\subsection{Dataset Statistics and Comparisons}
	In this part, we analyze the proposed ReDWeb-S dataset from several statistical aspects and also conduct a comparison between ReDWeb-S and other existing RGB-D SOD datasets, as shown in Table~\ref{dataCmpTab}.
	
	\textbf{Image Numbers:} This is a very important factor for modern data-hungry deep learning-based SOD models. From Table~\ref{dataCmpTab}, we can see that previous datasets have at most no more than 2,000 images and most of them have less than 1,000 images. Our ReDWeb-S dataset has a more than 1,000 increase and becomes the largest RGB-D SOD benchmark dataset.
	
	\textbf{Scene Statistics:} Rich scene and object categories are beneficial for both promoting and evaluating the generalization ability of SOD models. To evaluate the scene richness, we use a state-of-the-art ResNet-50 \cite{he2016resnet} based scene classification model pretrained on the Places365 \cite{zhou2017places} dataset to conduct scene categorization for each image in each dataset. Finally, we count the total scene class number for each dataset and report them in Table~\ref{dataCmpTab}. We can observe that our ReDWeb-S dataset has the most scene class number among all the nine datasets and it is much larger than those of most of the others. Similarly, we also evaluate the object richness of these datasets by conducting object classification. Specifically, we adopt a state-of-the-art ResNeXt-101 \cite{Xie2017resnext} based image classification model \cite{Touvron2019Fixing} to classify each image into one of the 1000-class object labels defined in ImageNet \cite{Deng2009ImageNet}. One thing to notice is that ImageNet labels do not include the ``person" class, which is very commonly seen in RGB-D SOD datasets. Therefore, we first manually pick out images whose foreground objects are persons and perform object classification on the remaining images. At last, we report the total object class number of each dataset in Table~\ref{dataCmpTab}. The results show that our dataset has the most diversiform objects. To have a deeper look at the scene and object diversities of our ReDWeb-S dataset, we also show its distributions of the top 60\% scene and object categories in Figure~\ref{fig:scene_obj_stat}. We can see that different scene categories have an approximately smooth distribution. However, for the object class distribution, we observe that nearly 20\% images of our ReDWeb-S dataset belong to the ``person" category, which is a dominant proportion in the distribution histogram. This is reasonable since ``person" is often the leading role in real-life photos and we also observe similar phenomena in other RGB-D SOD datasets, such as NJUD \cite{ju2014njud}, SSD \cite{zhu2017ssd}, and STERE \cite{niu2012stere}.
	
	\begin{figure*}[!t]
		\centering
		\subfigure[RGC: RGB global contrast]{
			\label{fig:GCICfig:a} 
			\includegraphics[width=0.24\linewidth]{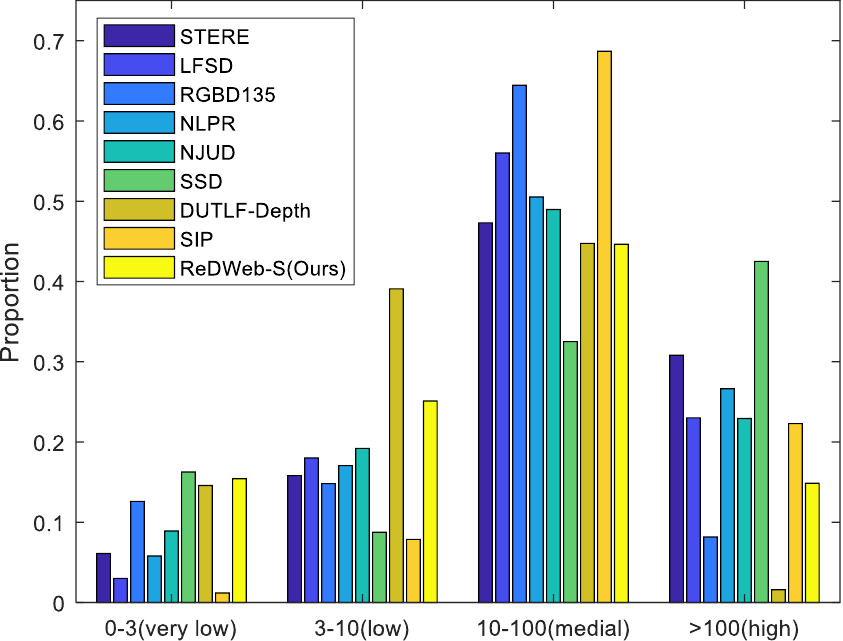}}
		\subfigure[DGC: Depth global contrast]{
			\label{fig:GCICfig:b} 
			\includegraphics[width=0.24\linewidth]{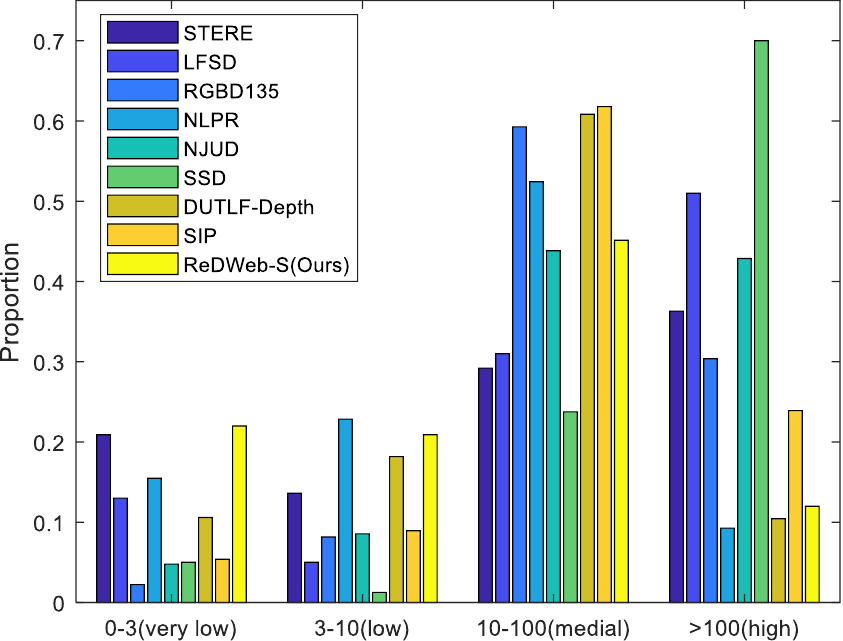}}
		\subfigure[RIC: RGB interior contrast]{
			\label{fig:GCICfig:c} 
			\includegraphics[width=0.24\linewidth]{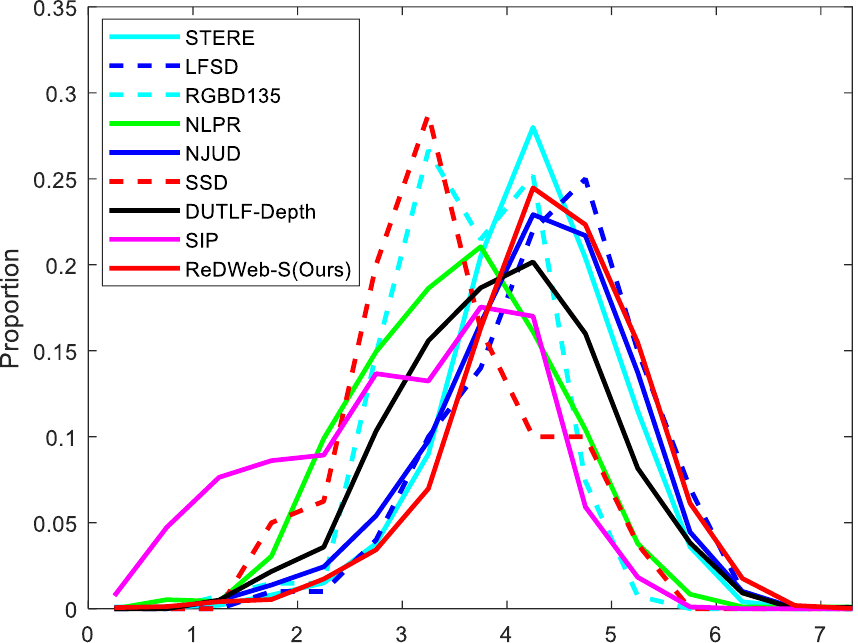}}
		\subfigure[DIC: Depth interior contrast]{
			\label{fig:GCICfig:d} 
			\includegraphics[width=0.24\linewidth]{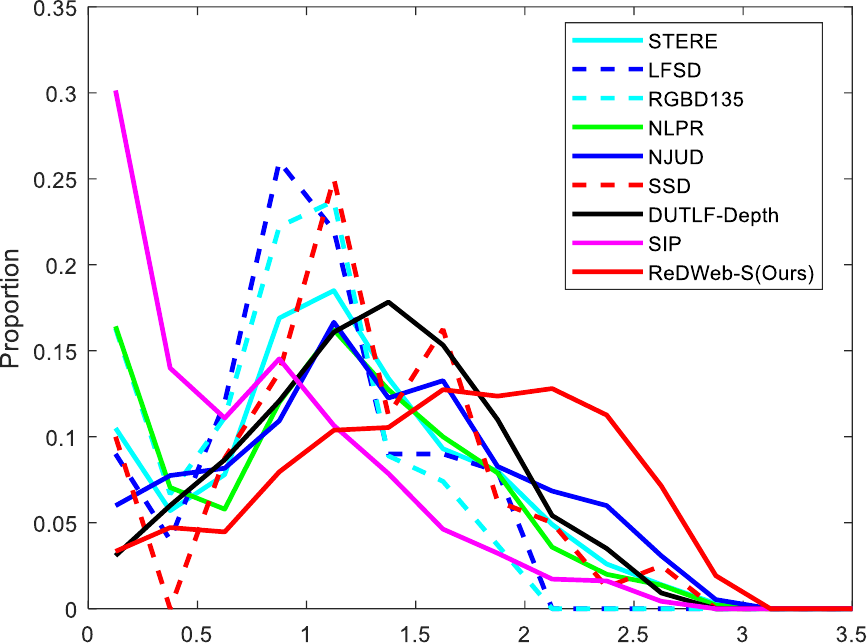}}
		\caption{\textbf{Comparison of nine RGB-D SOD dataset in terms of the distributions of global contrast and interior contrast.}}
		\label{fig:GCICfig} 
		\vspace{-0.3cm}
	\end{figure*}
	
	\begin{figure*}[!t]
		\graphicspath{{Figures/center bias/}}
		\centering
		\includegraphics[width=0.9\linewidth]{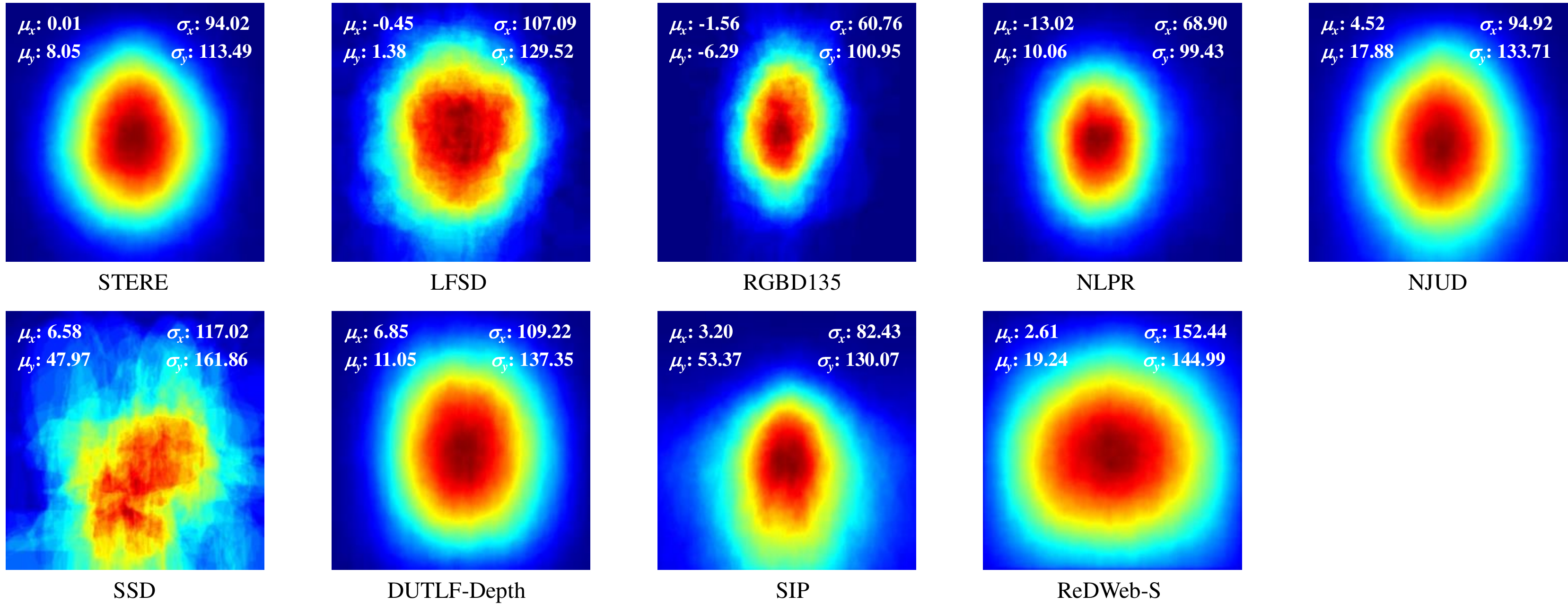}
		\caption{\textbf{Comparison of the average annotation maps for nine RGB-D SOD benchmark datasets.} We also use a 2D Gaussian distribution to fit each map and mark the corresponding coordinates of the  center point ($\mu_x$ and $\mu_y$) and the standard deviations ($\sigma_x$ and $\sigma_y$). }
		\label{fig:center_bias}
		\vspace{-0.3cm}
	\end{figure*}
	
	\textbf{Depth Map Quality:} Since depth maps provide very important complementary cues for saliency detection, their quality is also of significant importance. Depth maps with higher quality can supply more accurate guidance information for RGB-D SOD. Hence, we evaluate the depth map quality (DQ) based on the bad point rate (BPR) proposed in \cite{xiang2015esbm}. BPR is a state-of-the-art no-reference depth assessment metric and is calculated as the proportion of mismatched pixels between depth edges and texture edges. We define the DQ score as $1-BPR$, which is the matching accuracy. From Table~\ref{dataCmpTab} we can see that SIP has the best depth quality while ReDWeb-S ranks the second. We also observe that early datasets such as STERE and LFSD have the worst depth quality since they use the oldest depth estimation methods.
	
	\textbf{Global Contrast:} Since global contrast can help localize salient objects, it can be used to assess the challenging of each dataset. We first evaluate the global contrast for the RGB modality, denoted as ``RGC" in Table~\ref{dataCmpTab}. We follow \cite{fan2018soc} to compute the $\chi ^2$ distance between the RGB color histograms of foreground and background regions for each image as the measurement of RGC. Finally, we report its average value for each dataset in Table~\ref{dataCmpTab}. Similarly, we also report the measurements of the global contrast for the depth modality (denoted as ``DGC") by computing the $\chi ^2$ distance between depth histograms. The results demonstrate that ReDWeb-S has a relatively small global contrast. We also show the detailed distributions of RGC and DGC in Figure~\ref{fig:GCICfig}. Since the data ranges of the computed RGC and DGC are very large, we coarsely divide them into four scopes, \ie very low, low, medial, and high. We observe that our proposed ReDWeb-S dataset mainly have low and medial RGC, and very low, low, medial DGC. These results clearly demonstrate its challenging for RGB-D SOD.
	
	\textbf{Interior Contrast:} We refer to ``interior contrast" as the contrast within the salient object of each image. A small value means the different parts of a salient object have homogenous appearance, thus making it easier to uniformly highlight the whole object. On the contrary, a large value indicates the salient object is more sophisticated and harder to detect. We evaluate the interior contrast by computing the information entropy of the normalized color and depth histograms of foreground regions for the RGB and depth modality, respectively. The average RGB interior contrast (RIC) and depth interior contrast (DIC) of each dataset are reported in Table~\ref{dataCmpTab}. We find that ReDWeb-S achieves the largest average RIC and DIC. We also draw the curves of the distributions of RIC and DIC for all the datasets. Figure~\ref{fig:GCICfig:c} shows that ReDWeb-S and NJUD both have the most images with large RIC. Figure~\ref{fig:GCICfig:d} indicates ReDWeb-S has much more images with high DIC than other datasets. These observations further verify the difficulty of our proposed ReDWeb-S dataset.
	
	\begin{figure*}[!t]
		\graphicspath{{Figures/dataset_examp/}}
		\centering
		\includegraphics[width=1\linewidth]{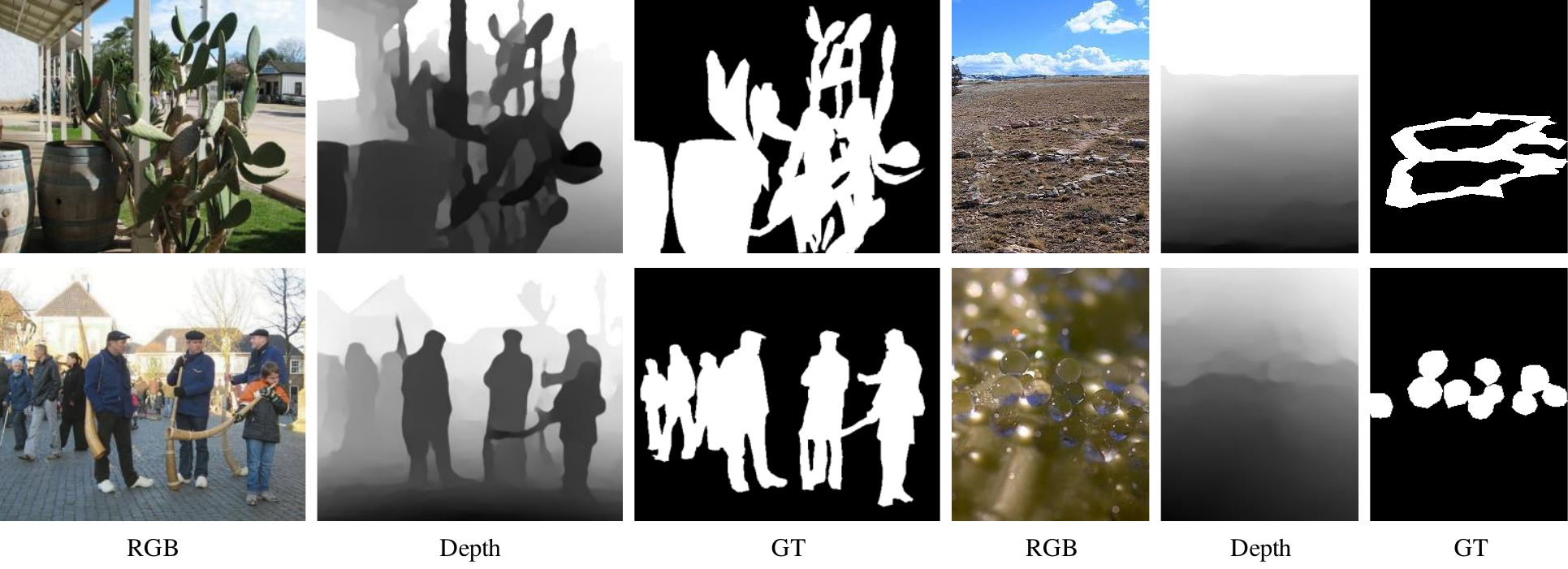}
		\caption{\textbf{Example images of our proposed ReDWeb-S dataset.}}
		\label{fig:data_examp}
		\vspace{-0.3cm}
	\end{figure*}
	
	\begin{figure}[h]
		\graphicspath{{Figures/OS/}}
		\centering
		\includegraphics[width=0.6\linewidth]{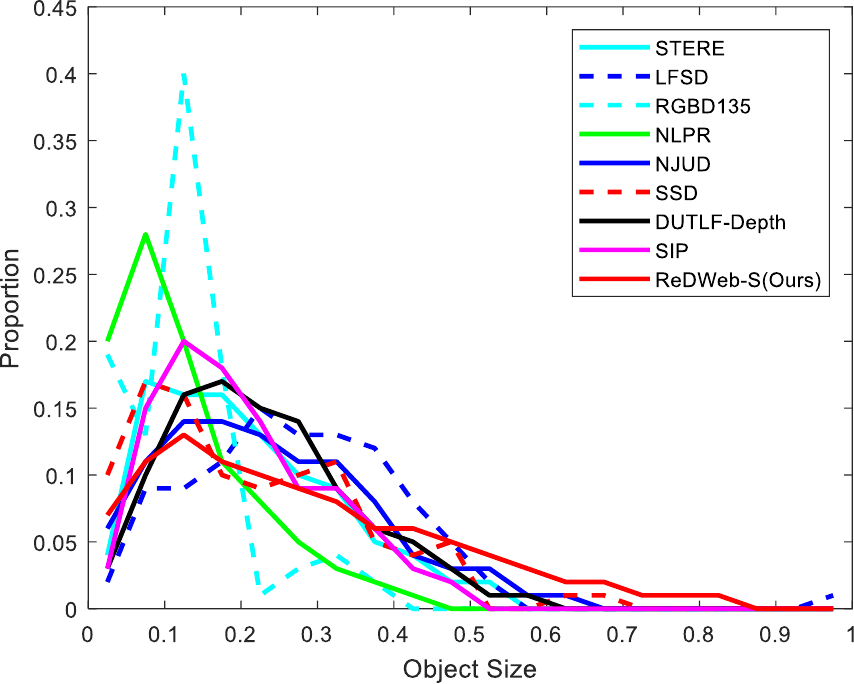}
		\caption{\textbf{Comparison of the distribution of object size (OS) for nine datasets.}}
		\label{fig:object_size}
		\vspace{-0.3cm}
	\end{figure}

	\textbf{Center Bias:} Center bias is a well-known prior knowledge for saliency detection since people usually put target objects in the middle of their photos. We follow previous works (\eg, \cite{borji2015sodbenchmark}) to draw the average annotation map (AAM), \ie the average of ground-truth annotations of all images, on each dataset to illustrate their center biases in Figure~\ref{fig:center_bias}. We resize each AAM to $256\times 256$ pixels to ease the comparison. For quantitatively analyzing the degree of center bias of each dataset, we propose to normalize each AAM and then use a 2D Gaussian distribution to fit it since most of them clearly show similar patterns with 2D Gaussian surfaces. Then, we mark the coordinates of the center point ($\mu_x$ and $\mu_y$) and the standard deviations ($\sigma_x$ and $\sigma_y$) along the width and the height direction on each AAM in Figure~\ref{fig:center_bias}. For $\mu_x$ and $\mu_y$, we report their offsets to the map center to ease understanding. Usually, larger center point offsets and standard deviations mean that one AAM deviates its center more, thus this dataset has less center bias. Among them, standard deviations are more important since the center points of most AAMs are close to the map centers and large standard deviations indicate salient objects spread in a large range on this dataset. From Figure~\ref{fig:center_bias}, we observe that ReDWeb-S has moderate center point offsets and the largest standard deviations, demonstrating it shows less center bias. We also compute the average of $\sigma_x$ and $\sigma_y$ as a center bias index (CBI) to comprehensively assess the degree of center bias for each dataset and report them in Table~\ref{dataCmpTab}. The comparison shows that ReDWeb-S achieves the largest CBI.
	
	
	\textbf{Object Size:} The sizes of salient objects also play an important role in SOD since usually both very large or small salient objects are difficult to detect. We compute the normalized object size for each image and draw its distribution for each dataset in Figure~\ref{fig:object_size}. It shows that most previous datasets usually have more small objects while ReDWeb-S has more large ones. The former is difficult to locate while it is hard to completely highlight the latter. We also report the average OS of each dataset in Table~\ref{dataCmpTab}. The results show that ReDWeb-S and LFSD achieve the largest average object size.
	
	Some example images of the ReDWeb-S dataset are given in Figure~\ref{fig:data_examp}. We can see that the depth maps are of high quality and there are various challenging scenarios, such as complex foreground and backgrounds, low-contrast images, and transparent objects.
	
	\section{Proposed Selective Mutual Attention and Contrast Model}
\begin{figure}[!t]
		\graphicspath{{Figures/SMAC/}}
		\centering
		\includegraphics[width=1\linewidth]{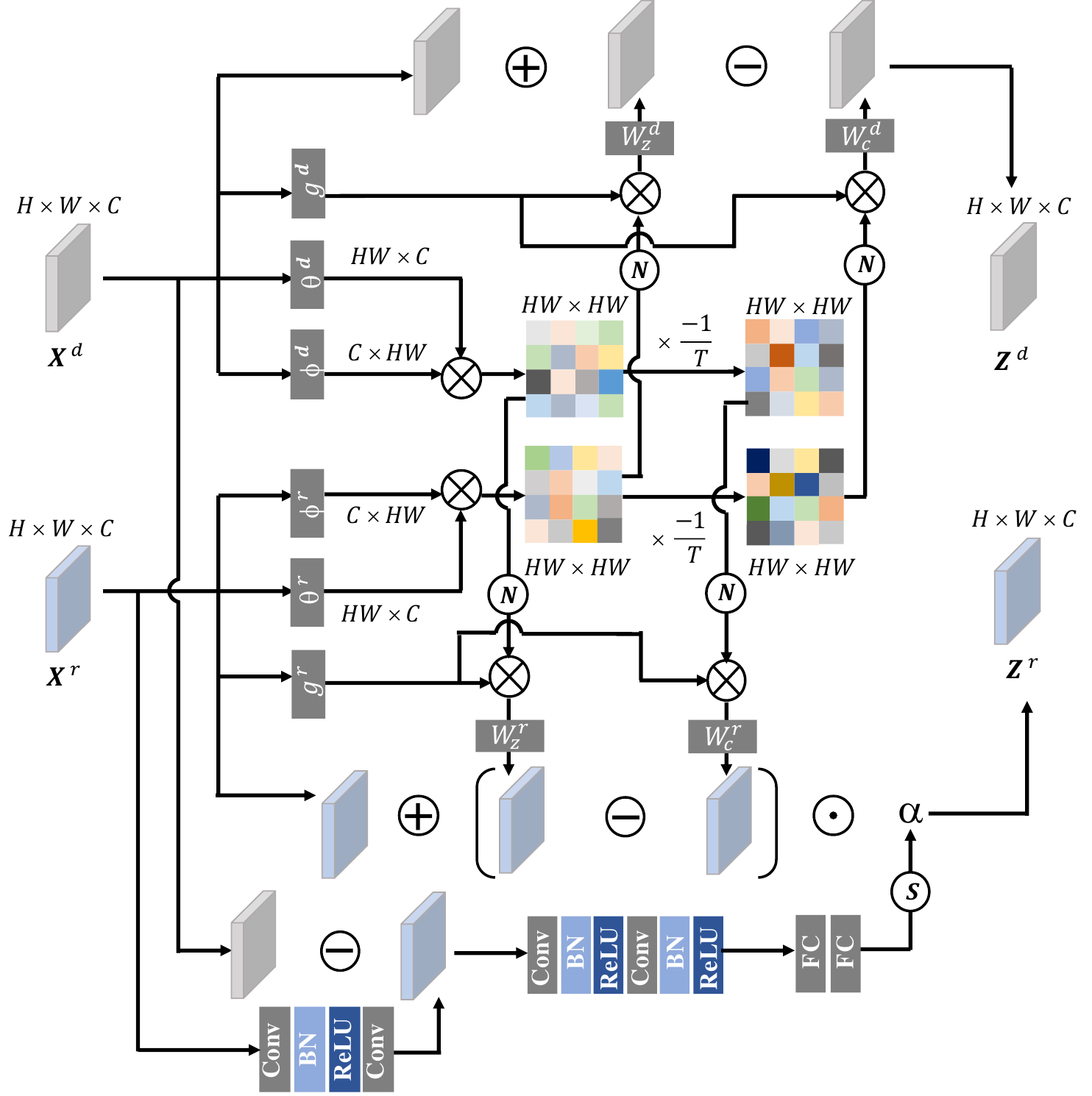}
		\caption{\textbf{Network architecture of the proposed SMAC module.} In the figure, $\otimes$, $\oplus$, $\ominus$, and $\odot$, indicate the matrix multiplication, addition, subtraction, and multiplication, respectively. \begin{normalsize}{\textcircled{\footnotesize{\emph{N}}}}\end{normalsize} and \begin{normalsize}{\textcircled{\footnotesize{\emph{S}}}}\end{normalsize} mean the Softmax normalization and the Sigmoid activation, respectively.}
		\label{fig:SMAC}
		\vspace{-0.3cm}
	\end{figure}

	In this section, we elaborate on the proposed Selective Mutual Attention and Contrast (SMAC) module for fusing multi-modal information in RGB-D SOD. It is built based on the NL module \cite{wang2018nonlocal} and the whole network architecture is shown in Figure~\ref{fig:SMAC}. We first briefly review the NL module and then go into our SMAC module.
	
	\subsection{Reviewing the NL module}
	
	Considering a feature map $\bm{X}\in{\mathbb{R}^{H\times{W\times C}}}$, where $H$, $W$, and $C$ represent its height, width, and channel number, respectively, the NL module first embeds $\bm{X}$ into three feature spaces with $C'$ channels:
	\begin{equation} \label{NL_embeding}
	\theta(\bm{X})=\bm{X}W_{\theta},\phi(\bm{X})=\bm{X}W_{\phi},g(\bm{X})=\bm{X}W_{g},
	\end{equation}
	where $W_{\theta}$, $W_{\phi}$, and $W_{g}\in{\mathbb{R}^{C\times{C'}}}$ are the embedding weights in the query, key, and value spaces, respectively. They can be implemented using $1\times 1$ convolutional (Conv) layers.
	
	Next, a similarity (or affinity) function $f$ is adopted using $\theta$ and $\phi$ embeddings as inputs. In \cite{wang2018nonlocal}, the authors have proposed several forms for the function $f$. Here we adopt the most widely used dot product function, \ie
	\begin{equation} \label{NL_aff}
	f(\bm{X})=\theta(\bm{X})\phi(\bm{X})^{\top},
	\end{equation}
	where $f(\bm{X})\in{\mathbb{R}^{HW\times{HW}}}$. In $f(\bm{X})$, each element $f_{i,j}$ represents the affinity between the query position $i$ and the key position $j$ in $\bm{X}$.
	
	Subsequently, the NL module adopts normalization along the second dimension to generate an attention weight matrix:
	\begin{equation} \label{NL_att}
	A(\bm{X})=softmax(f(\bm{X})),
	\end{equation}
	where each row $A_i$ indicates the normalized attention of all key positions respect to the $i^{th}$ query position. Then, the values in $g$ are aggregated by weighted sum:
	\begin{equation} \label{NL_attending}
	\bm{Y}=A(\bm{X})g(\bm{X}),
	\end{equation}
	where $\bm{Y}\in{\mathbb{R}^{HW\times{C'}}}$ is an attentive feature. By decomposing the computation for each query position, we have:
	\begin{equation} \label{NL_decom}
	\bm{Y}_i=\sum_{j=1}^{HW}A_{i,j}\cdot g_j.
	\end{equation}
	We can see that for each position in $\bm{Y}$, the obtained attentive feature is an aggregation of features at all positions. Thus $\bm{Y}$ incorporates long-range relations within the whole feature map $\bm{X}$.
	
	Finally, $\bm{Y}$ is first reshaped to the shape of $H\times{W\times C'}$, then the NL module learns a residual representation based on it to improve the original feature $\bm{X}$ and obtain a final output $\bm{Z}$:
	\begin{equation} \label{NL_output}
	\bm{Z}=\bm{X}+\bm{Y}W_z,
	\end{equation}
	where $W_{z}\in{\mathbb{R}^{C'\times{C}}}$ is the weight of a $1\times 1$ Conv layer for projecting the attentive feature back to the original feature space.

	\subsection{Mutual Attention}
	The obtaining of the attentive feature $\bm{Y}$ in the NL module can be rewritten as:
	\begin{equation} \label{NL_att2}
	\bm{Y}=softmax(\bm{X}W_{\theta}W_{\phi}^{\top}\bm{X}^{\top})\bm{X}W_{g}.
	\end{equation}
	We can see that it is a trilinear transform of the original feature $\bm{X}$ itself. Thus, it belongs to the self-attention category. We argue that the effectiveness of such a transform is highly limited by the quality of the original feature. If the original feature is of low-quality, the non-local attention can only attend to regions with self-similarity but without extra informative contexts. As a result, very limited feature promotion and performance gain can be achieved (see the experimental results in Section~\ref{sec:ablation}). For multi-modal tasks, such as RGB-D SOD, we can leverage the attention of different modalities to introduce context complementarity.
	
	In this paper, we propose using Mutual Attention (MA) for RGB-D SOD. Imaging we have two feature maps $\bm{X}^r,\bm{X}^d\in{\mathbb{R}^{H\times{W\times C}}}$ coming from the RGB and the depth modality, respectively, we first follow the NL module to embed them into the query, key spaces and obtain their attention matrixes:
	\begin{equation} \label{MA_att}
	\begin{split}
	A^r(\bm{X}^r)&=softmax(\theta^r(\bm{X}^r)\phi^r(\bm{X}^r)^{\top}),\\
	A^d(\bm{X}^d)&=softmax(\theta^d(\bm{X}^d)\phi^d(\bm{X}^d)^{\top}).
	\end{split}
	\end{equation}
	Then, we fuse multi-modal information by mixing the value modality and the attention modality up:
	\begin{equation} \label{MA_attending}
	\begin{split}
	\bm{Y}^r&=A^d(\bm{X}^d)g^r(\bm{X}^r),\\
	\bm{Y}^d&=A^r(\bm{X}^r)g^d(\bm{X}^d).
	\end{split}
	\end{equation}
	Here the two modalities provide attention for each other, thus we refer to this attention scheme as \emph{mutual attention}. It propagates cross-modal long-range contextual dependencies, which is a novel way for fusing multi-modal information. 
	
	By omitting the embedding weights and the Softmax normalization, we can rewrite the position-wise attentive features for the mutual attention as:
	\begin{equation} \label{MA_attending_rewrite}
	\begin{split}
	\bm{Y}^r_i&=\sum_{j=1}^{HW}<\bm{X}^d_i,\bm{X}^d_j>\bm{X}^r_j,\\
	\bm{Y}^d_i&=\sum_{j=1}^{HW}<\bm{X}^r_i,\bm{X}^r_j>\bm{X}^d_j,
	\end{split}
	\end{equation}
	where $<,>$ denotes the inner product of two feature vectors, and $\bm{X}^*_i$ means the feature vector at position $i$. Reviewing previous widely used modality fusion methods such as summation, multiplication, and concatenation, they only involve point-to-point low-order fusion. We observe from \eqref{MA_attending_rewrite} that our mutual attention module introduces high-order and trilinear interactions between $\bm{X}^r$ and $\bm{X}^d$, thus being able to explore more complex cross-modal information interaction.
	
	\subsection{Incorporating the Contrast Mechanism}
	The above attention mines spatial affinity via the feature inner product, thus integrating contexts with similar features from both views. Considering the effectiveness of the widely used contrast mechanism in SOD, which devotes to find the difference between salient regions and backgrounds, in this paper we also propose to incorporate the contrast mechanism. Reviewing that \eqref{NL_aff} computes the spatial affinity, we can easily obtain spatial dissimilarity by taking its opposite and then calculate a contrast attention:
	\begin{equation} \label{ctr_att}
	\mathcal{C}(\bm{X})=softmax(\frac{-f(\bm{X})}{T}),
	\end{equation}
	where the temperature $T$ is a learnable parameter to help learn an adaptive distribution for the contrast attention. Then, we can use this attention to aggregate features from contrastive regions and compute the feature difference as contrast. When integrating it with the proposed mutual attention, we can obtain a unified mutual attention and contrast (MAC) model and compute the modality-specific outputs $\bm{Z}^r$ and $\bm{Z}^d$ as:
	\begin{equation} \label{mac_output_r}
	\bm{Z}^r=\bm{X^r}+A^d(\bm{X}^d)g^r(\bm{X}^r)W_z^r-\mathcal{C}^d(\bm{X}^d)g^r(\bm{X}^r)W_c^{r},
	\end{equation}
	\begin{equation} \label{mac_output_d}
	\bm{Z}^d=\bm{X^d}+A^r(\bm{X}^r)g^d(\bm{X}^d)W_z^d-\mathcal{C}^r(\bm{X}^r)g^d(\bm{X}^d)W_c^{d},
	\end{equation}
	where $W_c^*\in{\mathbb{R}^{C'\times{C}}}$ plays a similar role with $W_z^*$.
	
	\begin{figure*}[!t]
		\graphicspath{{Figures/Network/}}
		\centering
		\includegraphics[width=0.9\linewidth]{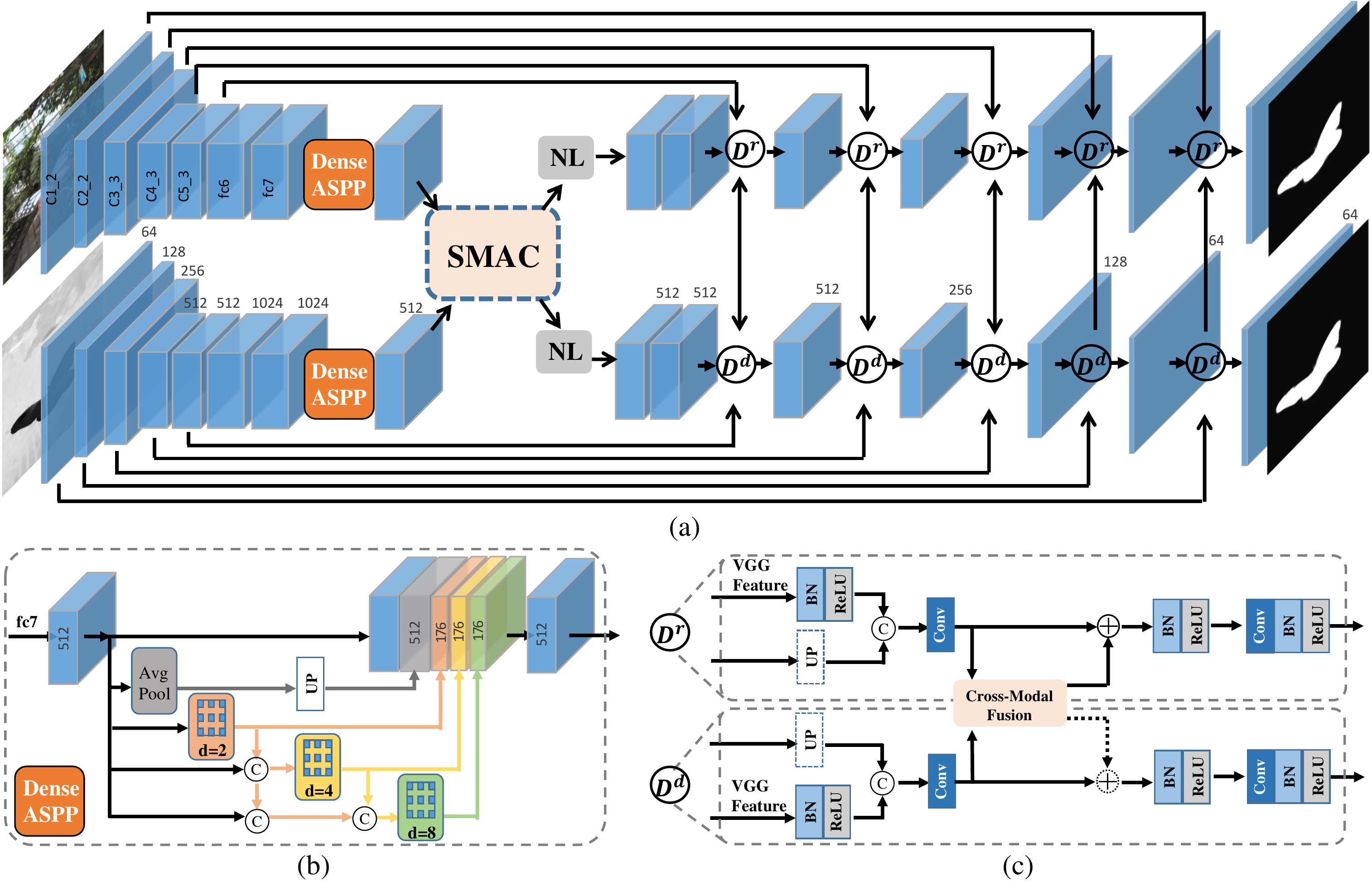}
		\caption{\textbf{Architecture of our proposed RGB-D SOD network.} (a) shows the main two-stream network. The skip-connected VGG layers are marked in the first stream by ``C*\_*'' and ``fc*''. The channel numbers of the feature maps are also marked in the second stream. ``NL" means the Non-local module \cite{wang2018nonlocal}. (b) shows the structure of our DenseASPP module. Some key channel numbers are also given. (c) shows the proposed decoder modules for the two streams. Here ``UP'' means upsampling with bilinear interpolation. ``Cross-Modal Fusion" can be either the proposed selective mutual attention module for the first three decoders or a simple concatenation based unidirectional fusion method for the last two decoders.}
		\label{network_fig}
		\vspace{-0.3cm}
	\end{figure*}
	
	\subsection{Selective Attention}
	The MAC model treats the information from the two modalities equally. However, considering that SOD using RGB data only has achieved very promising results yet \cite{zhao2019egnet,qin2019basnet,Zhou2020Interactive} and several RGB-D SOD benchmark datasets have low-quality depth maps, we regard the depth modality as complementary information and propose to reweight the depth cues using adaptive selective attention weights. In our previous work \cite{liu2020s2ma}, we proposed the pixel-wise selective attention for both modalities. However, the attention weights are inferred by simply concatenating both RGB and depth features. Such an implicit and pixel-wise learning method makes the training no easy and we found it hard to work for our new MAC model. In this work, we propose a novel method to explicitly infer image-wise selective attention, which is more effective for our new model and the attention weight can be used for multiple cross-modal fusion modules. Inspired by recent monocular depth estimation models, we argue that depth information can be roughly estimated by the RGB features of each image. If the estimation error is large, it suggests the original depth map is probably of low-quality. Thus, we can infer the selective attention from the estimation error and use it to weight the incorporated depth cues.
	
	Specifically, we first deploy two $1\times 1$ Conv layers with $C$ channels on the top of $\bm{X}^r$ to estimate $\bm{X}^d$. Then, the estimation error map can be computed by subtracting the estimated $\bm{X}^d$ from the real one. Next, two $1\times 1$ Conv layers with setting $stride=2$ are used to downsample the error map. We also reduce the channel numbers by setting them to 256 and 128, respectively. Batch normalization \cite{ioffe2015bn} and the ReLU activation function are used right after the first and the last two Conv layers, as shown in Figure~\ref{fig:SMAC}. Finally, two FC layers with 256 and 1 nodes are used to predict the final selective attention $\alpha$, with adopting the Sigmoid activation function. The whole process can be roughly represented as:
	\begin{equation} \label{select_att}
	\alpha=Sigmoid(FC(Conv(\bm{X}^d-Conv(\bm{X}^r)))).
	\end{equation}
	
	Then, we can use the selective attention to weight the mutual attention and the contrast terms in \eqref{mac_output_r} since they are induced from the depth modality and may suffer from the low-quality of depth maps:
	\begin{equation} \label{SMAC_att}
	\bm{Z}^r=\bm{X^r}+\alpha \cdot (A^d(\bm{X}^d)g^r(\bm{X}^r)W_z^r-\mathcal{C}^d(\bm{X}^d)g^r(\bm{X}^r)W_c^{r}).
	\end{equation}
	
	
	\section{RGB-D SOD Network}\label{sec:network}
	
	Based on the proposed SMAC module, we propose a novel deep model for RGB-D SOD. As shown in Figure~\ref{network_fig}(a), our model is based on a two-stream CNN, and each of them is responsible for generating the saliency map from the input modality based on the UNet \cite{ronneberger2015unet} architecture.
	
	Specifically, we share the same network structure for the two encoders and adopt the VGG-16 network \cite{simonyan2014vgg} as the backbone. We follow \cite{liu2018picanet} to slightly modify its network structure as follows. First, we change the pooling strides of the pool4 and pool5 layers to 1 and set the dilation rates \cite{chen2017deeplab} of the conv5 block to 2. Second, we transform the fc6 layer to a $3\times3$ Conv layer with 1024 channels and set its dilation rate to 12. Similarly, we turn the fc7 layer into a $1\times1$ Conv layer with 1024 channels. As such, the encoder network becomes a fully convolutional network \cite{long2015fcn} with the output stride of 8, thus preserving large spatial resolutions for high-level feature maps.
	
	Next, we adopt DenseASPP \cite{yang2018denseaspp} modules on top of the two encoders for further enhancing their capability. DenseASPP introduces dense connections \cite{huang2017densenet} to the ASPP \cite{chen2017deeplab} module and therefore covers dense feature scales. Before adopting DenseASPP, we first use $1\times 1$ Conv layers to compress the two fc7 feature maps to 512 channels. Considering the specific training image size of our SOD network, we design three dilated Conv branches with dilation rates of 2, 4, and 8, respectively. All of them use $3\times 3$ Conv kernels and 176 channels. Following \cite{yang2018denseaspp}, dense connections are deployed within the three branches. To capture the global context, we additionally design a fourth branch that average pools the input feature map first and then upsamples the result to the original size. At last, we concatenate the original input feature map and the outputs of the four branches, and then
	compress them to 512 channels. The whole module architecture is shown in Figure~\ref{network_fig}(b).
	
	After the DenseASPP module, we take the output features of the RGB and depth streams as inputs and adopt the proposed SMAC module to perform cross-modal information interaction. Since the outputs $\bm{Z}^r$ and $\bm{Z}^d$ have leveraged context propagation and contrast inference from the cross-modal attention, their quality and discriminability have been improved. Hence, we further use an NL module for each of them to blend the received cross-modal cues, as shown in Figure~\ref{network_fig}(a).
	
	Next, we go into the decoder parts. We represent the following decoder modules of the two branches as \begin{large}{\textcircled{\scriptsize{$D^r$}}}\end{large} and \begin{large}{\textcircled{\scriptsize{$D^d$}}}\end{large}, respectively.
	As shown in Figure~\ref{network_fig}(c), for each decoder module, we first follow the UNet \cite{ronneberger2015unet} architecture to progressively fuse an intermediate encoder feature map with the previous decoder feature map. The used intermediate VGG features are the last Conv feature maps of the five blocks, which are marked in Figure~\ref{network_fig}(a). For encoder-decoder feature fusion, we simply concatenate them together and then adopt two Conv layers. To enhance cross-modal information interactions, we also deploy cross-modal fusion structures between \begin{large}{\textcircled{\scriptsize{$D^r$}}}\end{large} and \begin{large}{\textcircled{\scriptsize{$D^d$}}}\end{large}. Concretely, for the first three of them, we use the proposed selective mutual attention (SMA) modules. Here we do not incorporate contrast anymore to save computational costs. For the last two decoder modules, we choose to not adopt the SMA modules since they are computationally prohibitive for large feature maps. As an alternative, we simply concatenate cross-modal decoder features and then use another Conv layer to learn a residual fusion signal for the RGB branch. Note that here the fusion is unidirectional since we consider RGB information as the main character. The selective attention is also adopted for this residual fusion.
	
	Each Conv layer in our decoder part uses $3\times 3$ kernels and is followed by a BN \cite{ioffe2015bn} layer and the ReLU activation function. We set the output channel number of each decoder module to be the same as that of the next skip-connected VGG feature map, as marked in Figure~\ref{network_fig}(a). For the last three decoder modules, we upsample previous decoder feature maps since they have smaller spatial sizes than the skip-connected VGG feature maps. Hence, the decoder feature maps are progressively enlarged. Due to the third \begin{large}{\textcircled{\scriptsize{$D^r$}}}\end{large} and \begin{large}{\textcircled{\scriptsize{$D^d$}}}\end{large} have relatively large spatial sizes, \ie $\frac{1}{4}$ of the input size, we downsample the $\phi$ and $g$ feature maps in SMA by a factor of 2 using max-pooling to save computational costs. For saliency prediction, we adopt a $3\times 3$ Conv layer with 1 channel on the last decoder feature map and use the Sigmoid activation function to obtain the saliency map for each CNN stream.

\section{Experiments}
\subsection{Evaluation Metrics}
Following recent works, we adopt four metrics for a comprehensive evaluation. The first one is the F-measure score, which
treats SOD as a binary classification task and comprehensively considers both precision and recall. We follow most previous works to report the maximum F-measure (maxF) score under the optimal threshold. The second metric is the Structure-measure $S_m$ \cite{fan2017structure}. It considers structural similarities between the saliency maps and the ground truth from the perspectives of both region-level and object-level. The third metric is the Enhanced-alignment measure $E_\xi$ \cite{fan2018enhanced} which jointly evaluates both global statistics and local pixel matching. The last metric we use is the widely used Mean Absolute Error (MAE). It is computed as the average of the absolute difference between a saliency map and the corresponding ground truth.

\subsection{Implementation Details}
Recent works \cite{Piao2019dmra,liu2020s2ma,zhang2020ssf,piao2020a2dele} train deep RGB-D SOD models using images from three datasets, \ie 1,400 images of the NJUD dataset, 650 images of the NLPR dataset, and 800 images of the DUTLF-Depth dataset. We follow this criterion to train our network. We set the training and testing image size of our SMAC RGB-D SOD network as $256\times 256$. Random cropping and horizontal flipping are adopted for data augmentation. As for the former, we resize each training image and the corresponding depth map to $288\times 288$ pixels and then randomly crop $256\times 256$ image regions as the network inputs. For the depth stream of our network, we replicate each single-channel depth map thrice to fit its input layer. Considering that different datasets have different depth presentations, we preprocess their depth maps to a unified presentation, \ie small depth values indicate the pixels are close to the camera and vice verse. The depth maps are also normalized to the value range of [0,255] to ease the network training. Before feeding into the two-stream network, each image and the corresponding three-channel depth map are subtracted by their mean pixel values as preprocessing.

We adopt the cross-entropy loss to train both streams of our SMAC network. Following previous works, deep supervision is also adopted in each decoder module to facilitate the network training. 
By following \cite{liu2018picanet}, we empirically use 0.5, 0.5, 0.8, 0.8, and 1 as the loss weights of the five decoder modules of each stream. We train our SMAC network totally using 40,000 iterations with the stochastic gradient descent (SGD) with momentum algorithm, where the initial learning rate, weight decay, momentum, and batchsize are set to 0.01, 0.0005, 0.9, and 12, respectively. We decay the learning rate by dividing it by 10 at the $20,000^{th}$ and the $30,000^{th}$ training steps, respectively.

Our SMAC model is implemented using the Pytorch \cite{paszke2017automatic} package. A GTX 1080 Ti GPU is used for computing acceleration. When testing, we resize each image pair to $256\times 256$ pixels as the network inputs and use the outputted saliency map from the RGB stream as the final saliency map, without using any post-processing technique. Our SMAC network only takes 0.059 seconds for testing each image-pair.

\subsection{Component Analysis}\label{sec:ablation}
For a more comprehensive evaluation, we additionally add our ReDWeb-S dataset for training in the component analysis experiments. However, due to space limitation and the ease of comparison, we only report the comparison of different model settings on three challenging datasets, \ie ReDWeb-S, NJUD, and LFSD. The experimental results are shown in Table~\ref{ablationTab}.

\begin{table*} [t]
	\begin{center}
		\caption{\textbf{Component analysis on the effectiveness of the proposed SMAC RGB-D SOD model.} We first show the comparison among different model settings of gradually using the proposed model components. \blu{Blue} indicates the best performance among these settings. We also compare our model with some other methods to verify its effectiveness, as shown in rows VII to XII.}
		\label{ablationTab}
		\footnotesize
		\begin{tabular}{@{}L{0.3cm}|L{2.4cm}|C{0.8cm}C{0.8cm}C{0.8cm}C{0.8cm}|C{0.8cm}C{0.8cm}C{0.8cm}C{0.8cm}|C{0.8cm}C{0.8cm}C{0.8cm}C{0.8cm}@{}}
			\toprule
			\multirow{2}{*}{ID} & \multirow{2}{*}{Settings} & \multicolumn{4}{c|}{ReDWeb-S} & \multicolumn{4}{c|}{NJUD \cite{ju2014njud}} & \multicolumn{4}{c}{LFSD \cite{li2014lfsd}} \\ \cmidrule{3-14}
			&&  $S_m$    & maxF      & $E_\xi$   &   MAE     &  $S_m$    & maxF      & $E_\xi$   &   MAE     &  $S_m$    & maxF      & $E_\xi$   &   MAE 
			\\ \midrule
			\multicolumn{14}{c}{\textbf{Gradually Using Proposed Model Components}}
			\\ \midrule
			I & Baseline &
			0.772  & 0.760  & 0.833  & 0.109      & 0.883  & 0.873  & 0.921  & 0.055      & 0.834  & 0.819  & 0.877  & 0.088
			\\
			II & +MA &
			0.784  & 0.769  & 0.838  & 0.115      & 0.898  & 0.895  & 0.934  & 0.054      & 0.862  & 0.864  & 0.901  & 0.077
			\\
			III & +MAC &
			0.787  & 0.772  & 0.842  & 0.113      & 0.901  & 0.896  & 0.936  & 0.052      & 0.862  & 0.863  & 0.902  & 0.078
			\\
			IV & +MAC+NL &
			0.787  & 0.772  & 0.841  & 0.109      & 0.904  & 0.900  & 0.939  & 0.049      & 0.874  & 0.868  & 0.907  & 0.071
			\\
			V & +MAC+NL+CMD &
			0.797  & 0.785  & 0.853  & 0.107      & 0.905  & 0.902  & 0.939  & 0.048      & \blu{0.880 } & \blu{0.881 } & \blu{0.917 } & 0.066
			\\
			VI & +SMAC+NL+SCMD &
			\blu{0.801 } & \blu{0.790 } & \blu{0.857 } & \blu{0.098 }     & \blu{0.911 } & \blu{0.908 } & \blu{0.942 } & \blu{0.043 }     & 0.878  & 0.874  & 0.909  & \blu{0.064 }
			\\ \midrule
			\multicolumn{14}{c}{\textbf{Compare Row I and II with Using Self-Attention \cite{wang2018nonlocal}}}
			\\ \midrule
			VII & +SA \cite{wang2018nonlocal} &
			0.765 & 0.749 & 0.823 & 0.113	 & 0.886 & 0.878 & 0.925 & 0.054	 & 0.821 & 0.796 & 0.864 & 0.093
			\\ \midrule
			\multicolumn{14}{c}{\textbf{Compare Row IV with Using $S^2$MA \cite{liu2020s2ma}}}
			\\ \midrule
			VIII & +$S^2$MA \cite{liu2020s2ma} &
			0.786 	& 0.770	& 0.841	& 0.110 	& 0.901	& 0.896	& 0.936	& 0.050 	& 0.862	& 0.862	& 0.906	& 0.076 
			\\ \midrule
			\multicolumn{14}{c}{\textbf{Compare Row V with Using the Concatenation Based CMD in All Five Decoders}}
			\\ \midrule
			IX & +MAC+NL+CMD' &
			0.787 & 0.774 & 0.844 & 0.110   & 0.908 & 0.908 & 0.943 & 0.046 	& 0.876 & 0.881 & 0.916 & 0.066  
			\\ \midrule
			\multicolumn{14}{c}{\textbf{Compare Row VI with Using Naive Fusion Methods}}
			\\ \midrule
			X & +Concat\_Fusion &
			0.781  & 0.767  & 0.838  & 0.116  & 0.903  & 0.901  & 0.937  & 0.052  & 0.865  & 0.862  & 0.903  & 0.078
			\\
			XI & +Sum\_Fusion &
			0.785  & 0.771  & 0.839  & 0.114  & 0.899  & 0.897  & 0.933  & 0.052 & 0.868  & 0.868  & 0.903  & 0.078
			\\
			XII & +Mul\_Fusion &
			0.786  & 0.770  & 0.841  & 0.114  & 0.901  & 0.897  & 0.937  & 0.051  & 0.870  & 0.870  & 0.905  & 0.076
			\\ \bottomrule
		\end{tabular}
		\vspace{-0.4cm}
	\end{center}{}
\end{table*}

\begin{figure*}[t]
	\scriptsize
	\renewcommand{\tabcolsep}{2pt} 
	\renewcommand{\arraystretch}{1} 
	\centering
	\begin{tabular}{cccccccccccc}
		\makecell[c]{\includegraphics[width=0.08\linewidth,height=0.08\linewidth]{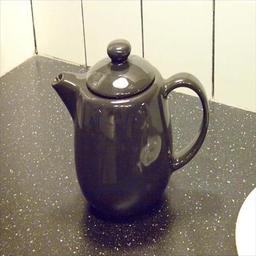}} &
		\makecell[c]{\includegraphics[width=0.08\linewidth,height=0.08\linewidth]{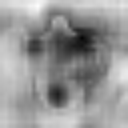}} &
		\makecell[c]{\includegraphics[width=0.08\linewidth,height=0.08\linewidth]{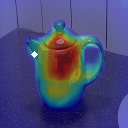}} &
		\makecell[c]{\includegraphics[width=0.08\linewidth,height=0.08\linewidth]{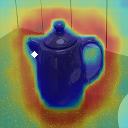}} &
		\makecell[c]{\includegraphics[width=0.08\linewidth,height=0.08\linewidth]{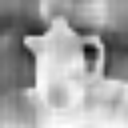}} &&&
		\makecell[c]{\includegraphics[width=0.08\linewidth,height=0.08\linewidth]{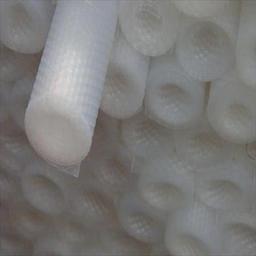}} &
		\makecell[c]{\includegraphics[width=0.08\linewidth,height=0.08\linewidth]{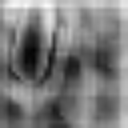}} &
		\makecell[c]{\includegraphics[width=0.08\linewidth,height=0.08\linewidth]{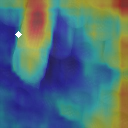}} &
		\makecell[c]{\includegraphics[width=0.08\linewidth,height=0.08\linewidth]{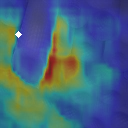}} &
		\makecell[c]{\includegraphics[width=0.08\linewidth,height=0.08\linewidth]{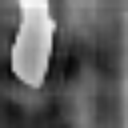}}
		\\
		RGB &
		$\bm{X}^r$ &
		$A^r(\bm{X}^r)$ &
		$\mathcal{C}^r(\bm{X}^r)$ &
		$\bm{Z}^r$ &&&
		RGB &
		$\bm{X}^r$ &
		$A^r(\bm{X}^r)$ &
		$\mathcal{C}^r(\bm{X}^r)$ &
		$\bm{Z}^r$
		\\
		\\
		\makecell[c]{\includegraphics[width=0.08\linewidth,height=0.08\linewidth]{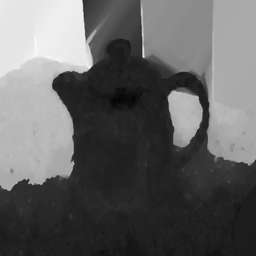}} &
		\makecell[c]{\includegraphics[width=0.08\linewidth,height=0.08\linewidth]{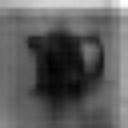}} &
		\makecell[c]{\includegraphics[width=0.08\linewidth,height=0.08\linewidth]{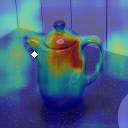}} &
		\makecell[c]{\includegraphics[width=0.08\linewidth,height=0.08\linewidth]{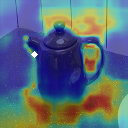}} &
		\makecell[c]{\includegraphics[width=0.08\linewidth,height=0.08\linewidth]{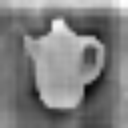}} &&&
		\makecell[c]{\includegraphics[width=0.08\linewidth,height=0.08\linewidth]{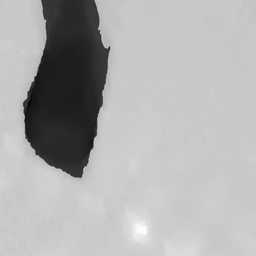}} &
		\makecell[c]{\includegraphics[width=0.08\linewidth,height=0.08\linewidth]{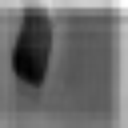}} &
		\makecell[c]{\includegraphics[width=0.08\linewidth,height=0.08\linewidth]{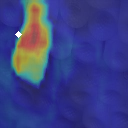}} &
		\makecell[c]{\includegraphics[width=0.08\linewidth,height=0.08\linewidth]{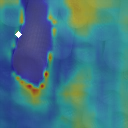}} &
		\makecell[c]{\includegraphics[width=0.08\linewidth,height=0.08\linewidth]{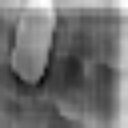}}
		\\
		Depth &
		$\bm{X}^d$ &
		$A^d(\bm{X}^d)$ &
		$\mathcal{C}^d(\bm{X}^d)$ &
		$\bm{Z}^d$ &&&
		Depth &
		$\bm{X}^d$ &
		$A^d(\bm{X}^d)$ &
		$\mathcal{C}^d(\bm{X}^d)$ &
		$\bm{Z}^d$
		\\
	\end{tabular}
	\caption{
		\textbf{Visualization of some attention maps and feature maps.} We show the feature maps ($\bm{X}^*$), the attention maps ($A^*(\bm{X}^*)$), the contrast attention maps ($\mathcal{C}^*(\bm{X}^*)$), and the output feature maps of the SMAC module ($\bm{Z}^*$) for the RGB and depth modalities in two image pairs. In each image, the white point indicates the query position.
	}
	\label{fig:att&fm}
	\vspace{-0.3cm}
\end{figure*}

\begin{figure}[t]
	\scriptsize
	\renewcommand{\tabcolsep}{1pt} 
	\renewcommand{\arraystretch}{1} 
	\centering
	\begin{tabular}{cccccc}
		\makecell[c]{\includegraphics[width=0.16\linewidth,height=0.16\linewidth]{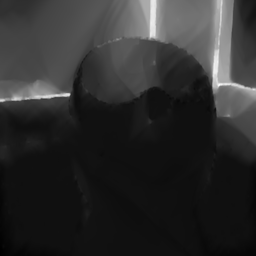}} &
		\makecell[c]{\includegraphics[width=0.16\linewidth,height=0.16\linewidth]{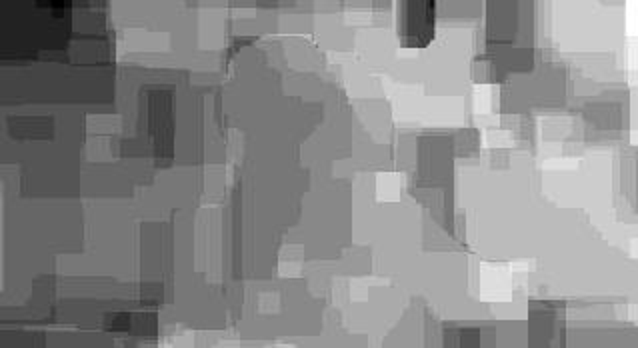}} &
		\makecell[c]{\includegraphics[width=0.16\linewidth,height=0.16\linewidth]{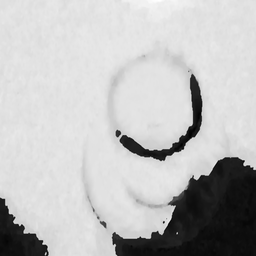}} &
		\makecell[c]{\includegraphics[width=0.16\linewidth,height=0.16\linewidth]{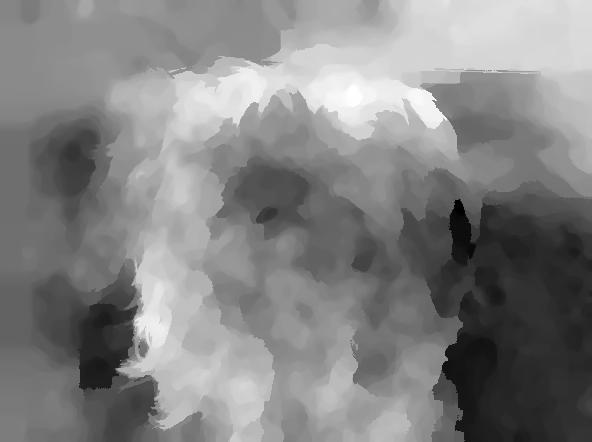}} &
		\makecell[c]{\includegraphics[width=0.16\linewidth,height=0.16\linewidth]{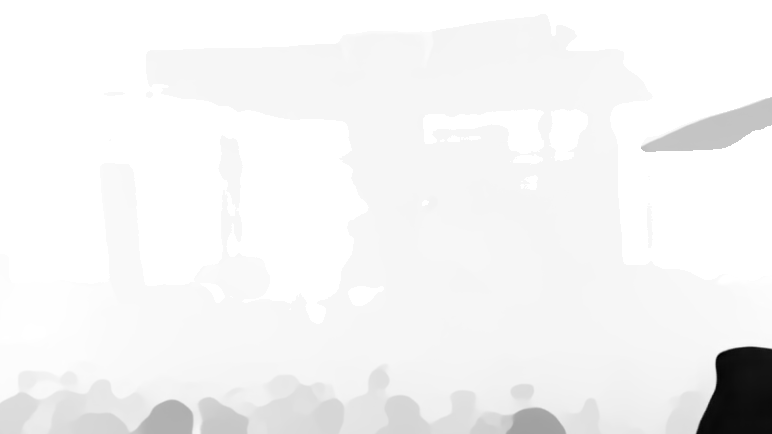}} &
		\makecell[c]{\includegraphics[width=0.16\linewidth,height=0.16\linewidth]{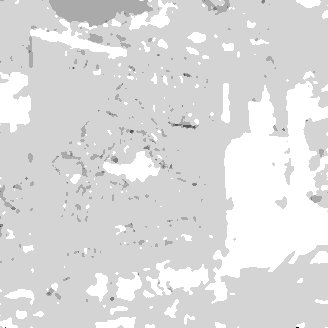}}
		\\
		\makecell[c]{\includegraphics[width=0.16\linewidth,height=0.16\linewidth]{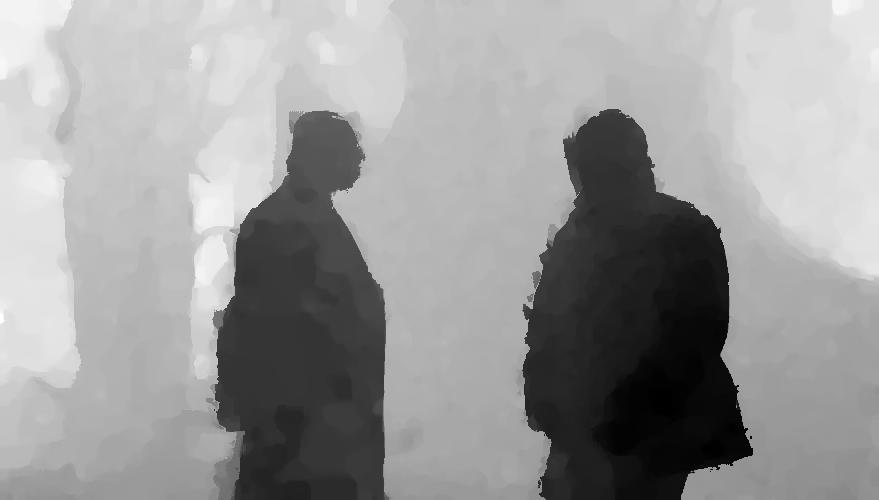}} &
		\makecell[c]{\includegraphics[width=0.16\linewidth,height=0.16\linewidth]{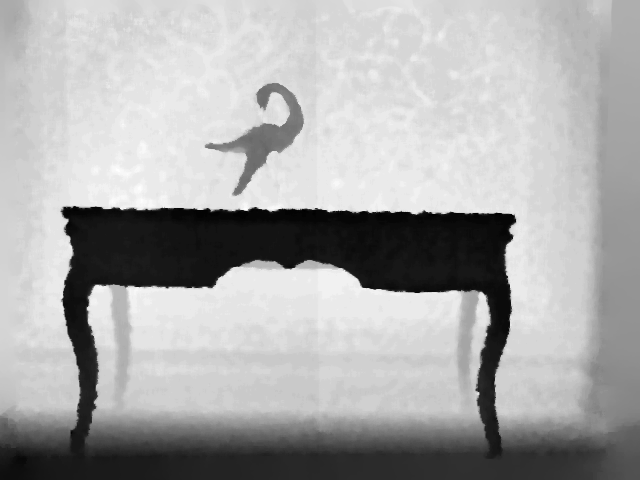}} &
		\makecell[c]{\includegraphics[width=0.16\linewidth,height=0.16\linewidth]{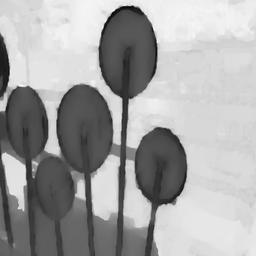}} &
		\makecell[c]{\includegraphics[width=0.16\linewidth,height=0.16\linewidth]{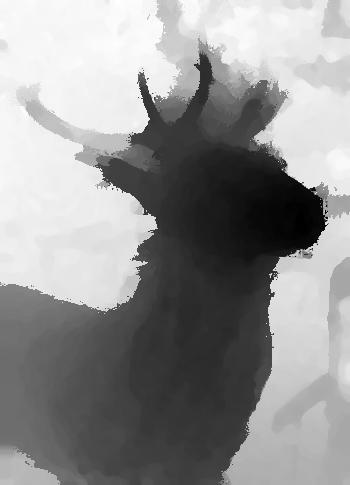}} &
		\makecell[c]{\includegraphics[width=0.16\linewidth,height=0.16\linewidth]{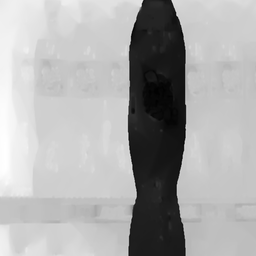}} &
		\makecell[c]{\includegraphics[width=0.16\linewidth,height=0.16\linewidth]{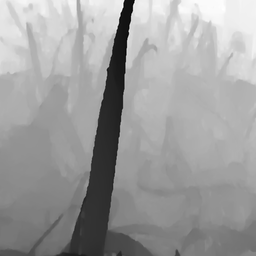}}
		\\
	\end{tabular}
	\caption{
		\textbf{\textbf{Example depth maps for the smallest (top row) and largest (bottom row) selective attention weights.}}
	}
	\label{fig:sel_att_img}
	\vspace{-0.3cm}
\end{figure}

\begin{table*} [t]
	\begin{center}
		\caption{\textbf{Quantitative comparison of our proposed model with other 10 state-of-the-art RGB-D SOD methods.} The comparison is conducted on 9 benchmark datasets in terms of 4 evaluation metrics. \red{Red} and \blu{blue} indicate the best and the second best performance, respectively.}
		\label{SOTATab}
		\footnotesize
		\begin{tabular}{@{}L{1.3cm}R{1cm}|C{0.65cm}C{0.8cm}C{0.8cm}C{0.8cm}C{0.8cm}C{0.8cm}C{0.8cm}C{0.8cm}C{0.8cm}C{0.8cm}|C{0.8cm}C{1.1cm}@{}}
			\toprule
			\multirow{2}{*}{Dataset} &\multirow{2}{*}{Metric} &DF &AFNet &CTMF &MMCI &PCF &TANet &CPFP &DMRA &D$^3$Net &$S^2$MA &SMAC &SMAC* 
			\\
			&&\cite{qu2017df} &\cite{wang2019afnet} &\cite{han2017ctmf} &\cite{chen2019mmci} &\cite{chen2018pcf} &\cite{chen2019tanet} &\cite{zhao2019cpfp} &\cite{Piao2019dmra} &\cite{fan2019d3net} &\cite{liu2020s2ma} &(Ours) &\begin{tiny}{(+ReDWeb-S)}\end{tiny} \\ \cmidrule{1-14}
			NJUD &$S_m$ $\uparrow$ &
			0.763  & 0.772  & 0.849  & 0.858  & 0.877  & 0.878  & 0.878  & 0.886  & 0.900  & 0.894  & \blu{0.903 } & \red{0.911 }
			\\
			&maxF $\uparrow$ &
			0.804  & 0.775  & 0.845  & 0.852  & 0.872  & 0.874  & 0.877  & 0.886  & 0.900  & 0.889  & \blu{0.896 } & \red{0.908 }
			\\
			&$E_\xi$ $\uparrow$ &
			0.864  & 0.853  & 0.913  & 0.915  & 0.924  & 0.925  & 0.923  & 0.927  & 0.939  & 0.930  & \blu{0.937 } & \red{0.942 }
			\\
			\cite{ju2014njud} &MAE $\downarrow$ &
			0.141  & 0.100  & 0.085  & 0.079  & 0.059  & 0.060  & 0.053  & 0.051  & 0.046  & 0.053  & \blu{0.044 } & \red{0.043 }
			\\ \midrule
			NLPR &$S_m$ $\uparrow$ &
			0.802  & 0.799  & 0.860  & 0.856  & 0.874  & 0.886  & 0.888  & 0.899  & 0.912  & 0.915  & \blu{0.922 } & \red{0.926 }
			\\
			&maxF $\uparrow$ &
			0.778  & 0.771  & 0.825  & 0.815  & 0.841  & 0.863  & 0.867  & 0.879  & 0.897  & 0.902  & \blu{0.904 } & \red{0.912 }
			\\
			&$E_\xi$ $\uparrow$ &
			0.880  & 0.879  & 0.929  & 0.913  & 0.925  & 0.941  & 0.932  & 0.947  & \blu{0.953 } & \blu{0.953 } & \blu{0.953 } & \red{0.960 }
			\\
			\cite{peng2014nlpr} &MAE $\downarrow$ &
			0.085  & 0.058  & 0.056  & 0.059  & 0.044  & 0.041  & 0.036  & 0.031  & 0.030  & 0.030  & \blu{0.027 } & \red{0.026 }
			\\ \midrule
			RGBD135 &$S_m$ $\uparrow$ &
			0.752  & 0.770  & 0.863  & 0.848  & 0.842  & 0.858  & 0.872  & 0.900  & 0.898  & \red{0.941 } & \blu{0.935 } & 0.923
			\\
			&maxF $\uparrow$ &
			0.766  & 0.729  & 0.844  & 0.822  & 0.804  & 0.827  & 0.846  & 0.888  & 0.885  & \red{0.935 } & \blu{0.928 } & 0.906
			\\
			&$E_\xi$ $\uparrow$ &
			0.870  & 0.881  & 0.932  & 0.928  & 0.893  & 0.910  & 0.923  & 0.943  & 0.946  & \red{0.973 } & \blu{0.972 } & 0.958
			\\
			\cite{cheng2014rgbd135} &MAE $\downarrow$ &
			0.093  & 0.068  & 0.055  & 0.065  & 0.049  & 0.046  & 0.038  & 0.030  & 0.031  & \blu{0.021 } & \red{0.020 } & 0.024
			\\ \midrule
			LFSD &$S_m$ $\uparrow$ &
			0.791  & 0.738  & 0.796  & 0.787  & 0.794  & 0.801  & 0.828  & 0.847  & 0.825  & 0.837  & \blu{0.875 } & \red{0.878 }
			\\
			&maxF $\uparrow$ &
			0.817  & 0.744  & 0.791  & 0.771  & 0.779  & 0.796  & 0.826  & 0.856  & 0.810  & 0.835  & \blu{0.870 } & \red{0.874 }
			\\
			&$E_\xi$ $\uparrow$ &
			0.865  & 0.815  & 0.865  & 0.839  & 0.835  & 0.847  & 0.872  & 0.900  & 0.862  & 0.873  & \red{0.911 } & \blu{0.909 }
			\\
			\cite{li2014lfsd} &MAE $\downarrow$ &
			0.138  & 0.133  & 0.119  & 0.132  & 0.112  & 0.111  & 0.088  & 0.075  & 0.095  & 0.094  & \red{0.063 } & \blu{0.064 }
			\\ \midrule
			STERE &$S_m$ $\uparrow$ &
			0.757  & 0.825  & 0.848  & 0.873  & 0.875  & 0.871  & 0.879  & 0.886  & 0.899  & 0.890  & \blu{0.905 } & \red{0.908 }
			\\
			&maxF $\uparrow$ &
			0.757  & 0.823  & 0.831  & 0.863  & 0.860  & 0.861  & 0.874  & 0.886  & 0.891  & 0.882  & \blu{0.897 } & \red{0.902 }
			\\
			&$E_\xi$ $\uparrow$ &
			0.847  & 0.887  & 0.912  & 0.927  & 0.925  & 0.923  & 0.925  & 0.938  & 0.938  & 0.932  & \blu{0.941 } & \red{0.943 }
			\\
			\cite{niu2012stere} &MAE $\downarrow$ &
			0.141  & 0.075  & 0.086  & 0.068  & 0.064  & 0.060  & 0.051  & 0.047  & 0.046  & 0.051  & \red{0.042 } & \blu{0.043 }
			\\ \midrule
			SSD &$S_m$ $\uparrow$ &
			0.747  & 0.714  & 0.776  & 0.813  & 0.841  & 0.839  & 0.807  & 0.857  & 0.857  & 0.868  & \blu{0.884 } & \red{0.890 }
			\\
			&maxF $\uparrow$ &
			0.735  & 0.687  & 0.729  & 0.781  & 0.807  & 0.810  & 0.766  & 0.844  & 0.834  & 0.848  & \blu{0.869 } & \red{0.876 }
			\\
			&$E_\xi$ $\uparrow$ &
			0.828  & 0.807  & 0.865  & 0.882  & 0.894  & 0.897  & 0.852  & 0.906  & 0.910  & 0.909  & \red{0.928 } & \blu{0.927 }
			\\
			\cite{zhu2017ssd} &MAE $\downarrow$ &
			0.142  & 0.118  & 0.099  & 0.082  & 0.062  & 0.063  & 0.082  & 0.058  & 0.058  & 0.052  & \red{0.044 } & \blu{0.045 }
			\\ \midrule
			DUTLF- &$S_m$ $\uparrow$ &
			0.736  & 0.702  & 0.831  & 0.791  & 0.801  & 0.808  & 0.818  & 0.889  & 0.850  & 0.903  & \red{0.926 } & \blu{0.921 }
			\\
			Depth &maxF $\uparrow$ &
			0.740  & 0.659  & 0.823  & 0.767  & 0.771  & 0.790  & 0.795  & 0.898  & 0.842  & 0.901  & \red{0.928 } & \blu{0.924 }
			\\
			&$E_\xi$ $\uparrow$ &
			0.823  & 0.796  & 0.899  & 0.859  & 0.856  & 0.861  & 0.859  & 0.933  & 0.889  & 0.937  & \red{0.956 } & \blu{0.950 }
			\\
			\cite{Piao2019dmra} &MAE $\downarrow$ &
			0.144  & 0.122  & 0.097  & 0.113  & 0.100  & 0.093  & 0.076  & 0.048  & 0.071  & 0.043  & \red{0.033 } & \blu{0.039 }
			\\ \midrule
			SIP &$S_m$ $\uparrow$ &
			0.653  & 0.720  & 0.716  & 0.833  & 0.842  & 0.835  & 0.850  & 0.806  & 0.860  & 0.872  & \blu{0.883 } & \red{0.895 }
			\\
			&maxF $\uparrow$ &
			0.657  & 0.712  & 0.694  & 0.818  & 0.838  & 0.830  & 0.851  & 0.821  & 0.861  & 0.877  & \blu{0.886 } & \red{0.895 }
			\\
			&$E_\xi$ $\uparrow$ &
			0.759  & 0.819  & 0.829  & 0.897  & 0.901  & 0.895  & 0.903  & 0.875  & 0.909  & 0.919  & \blu{0.925 } & \red{0.936 }
			\\
			\cite{fan2019d3net} &MAE $\downarrow$ &
			0.185  & 0.118  & 0.139  & 0.086  & 0.071  & 0.075  & 0.064  & 0.085  & 0.063  & 0.057  & \blu{0.049 } & \red{0.046 }
			\\ \midrule
			\multirow{4}{*}{ReDWeb-S} &$S_m$ $\uparrow$ &
			0.595  & 0.546  &0.641   & 0.660  & 0.655  & 0.656  & 0.685  & 0.592  & 0.689  & 0.711  & \blu{0.723 } & \red{0.801 }
			\\
			&maxF $\uparrow$ &
			0.579  & 0.549  &0.607   & 0.641  & 0.627  & 0.623  & 0.645  & 0.579  & 0.673  & 0.696  & \blu{0.718 } & \red{0.790 }
			\\
			&$E_\xi$ $\uparrow$ &
			0.683  & 0.693  &0.739   & 0.754  & 0.743  & 0.741  & 0.744  & 0.721  & 0.768  & 0.781  & \blu{0.801 } & \red{0.857 }
			\\
			&MAE $\downarrow$ &
			0.233  & 0.213  &0.204   & 0.176  & 0.166  & 0.165  & 0.142  & 0.188  & 0.149  & 0.139  & \blu{0.125 } & \red{0.098 }
			\\
			\bottomrule
		\end{tabular}
		\vspace{-0.3cm}
	\end{center}{}
\end{table*}

\textbf{Effectiveness of Different Model Components.} The first row in Table~\ref{ablationTab} denotes the baseline model, \ie adopting UNet and DenseASPP for the RGB data only. In the second and the third row we show the model performance of using MA and MAC modules, respectively, to incorporate attention-based cross-modal interaction. We can see that adopting MA largely improves the model performance, especially on the LFSD dataset. The MAC module can further moderately improve the results by incorporating the contrast mechanism on two out of the three datasets. Then, we add NL modules on top of the output feature maps of the MAC module to further blend the received cross-modal cues. The results are reported in row IV and we find that further using NL modules after the MAC module can continue bringing performance gains, especially on the LFSD dataset. Next, we use cross-modal decoders (CMD) by adopting MA modules in the first three decoders and the simple concatenation based fusion method in the last two, as discussed in Section~\ref{sec:network}. The results in row V demonstrate that fusing cross-modal features in decoder modules can also promote the modal capability, especially on the ReDWeb-S dataset. At last, we adopt the computed selective attention in MAC and CMD to weight the depth cues and report the results in row VI. We observe that using this strategy can lead to performance improvements on two out of the three datasets. Hence, we use this model setting as our final RGB-D SOD model.

To thoroughly understand the effectiveness of our proposed SMAC module, we show some visualization examples of the input feature maps $\bm{X}^*$, the attention maps $A^*(\bm{X}^*)$, the contrast attention maps $\mathcal{C}^*(\bm{X}^*)$, and the output feature maps $\bm{Z}^*$ of the SMAC module in Figure~\ref{fig:att&fm}, for both RGB and depth modalities. In the left example, the RGB image is more discriminative than the depth map, thus the RGB attention maps $A^r(\bm{X}^r)$ and $\mathcal{C}^r(\bm{X}^r)$ are better than the depth attention maps. An opposite situation is given in the right example. We can see that after adopting the SMAC module, the four feature maps are all improved and become more discriminative, thus can benefit the final saliency prediction. To see what the selective attention has learned, we show some example depth maps for the smallest and largest selective attention weights in Figure~\ref{fig:sel_att_img}. In the top row, we observe that small attention weights are mainly generated for low-quality depth maps, which are over-smoothing, or inaccurate, or indiscriminating for localizing salient objects. On the contrary, the bottom row shows that depth maps have clear boundaries and discriminability for segmenting salient objects can obtain large attention weights.

\begin{figure*}[t]
	\scriptsize
	\renewcommand{\tabcolsep}{1pt} 
	\renewcommand{\arraystretch}{1} 
	\centering
	\begin{tabular}{cccccccccccccccc}
		I &
		\makecell[c]{\includegraphics[width=0.061\linewidth]{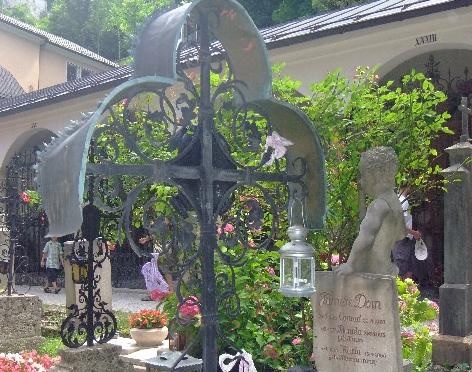}} &
		\makecell[c]{\includegraphics[width=0.061\linewidth]{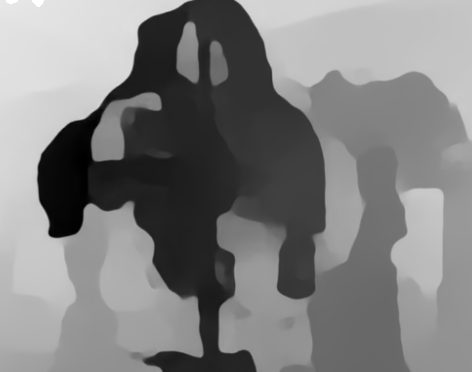}} &
		\makecell[c]{\includegraphics[width=0.061\linewidth]{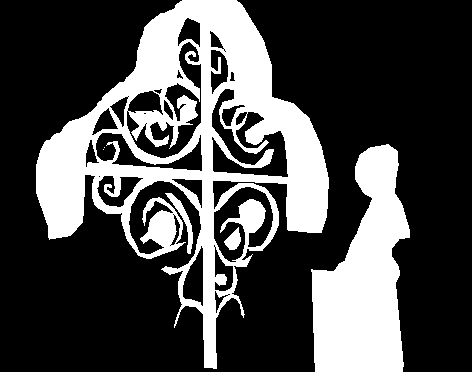}} &
		\makecell[c]{\includegraphics[width=0.061\linewidth]{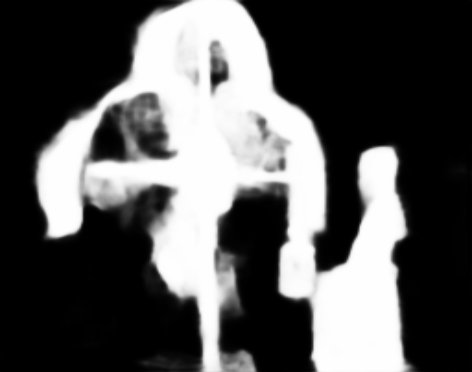}} &
		\makecell[c]{\includegraphics[width=0.061\linewidth]{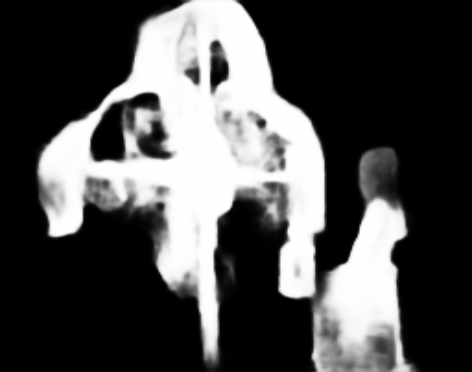}} &
		\makecell[c]{\includegraphics[width=0.061\linewidth]{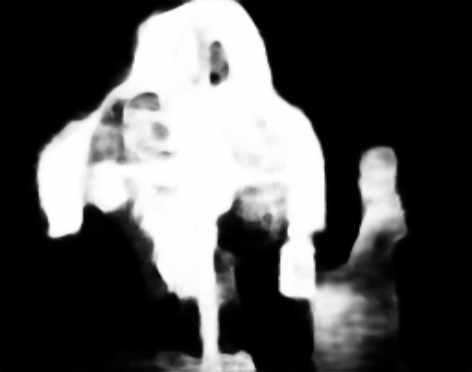}} &
		\makecell[c]{\includegraphics[width=0.061\linewidth]{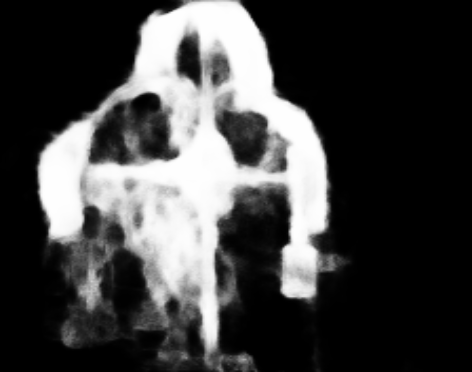}} &
		\makecell[c]{\includegraphics[width=0.061\linewidth]{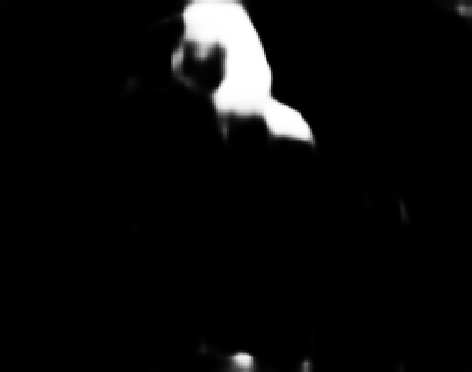}} &
		\makecell[c]{\includegraphics[width=0.061\linewidth]{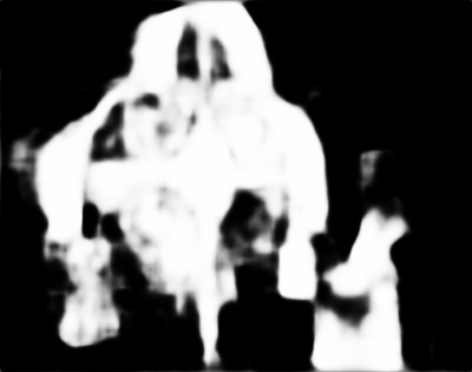}} &
		\makecell[c]{\includegraphics[width=0.061\linewidth]{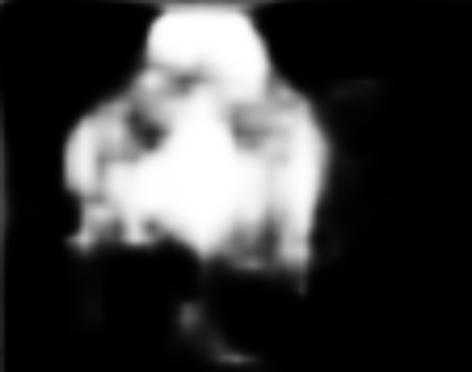}} &
		\makecell[c]{\includegraphics[width=0.061\linewidth]{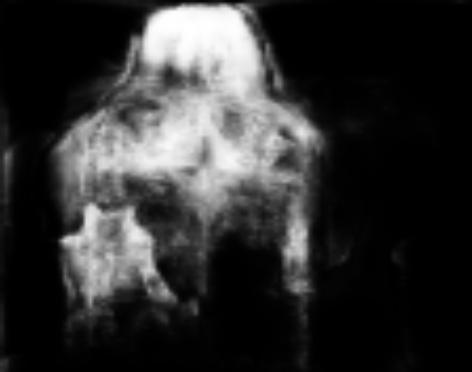}} &
		\makecell[c]{\includegraphics[width=0.061\linewidth]{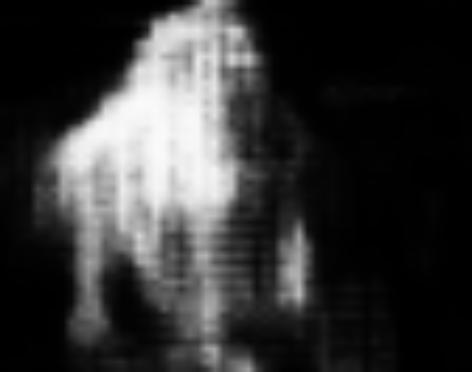}} &
		\makecell[c]{\includegraphics[width=0.061\linewidth]{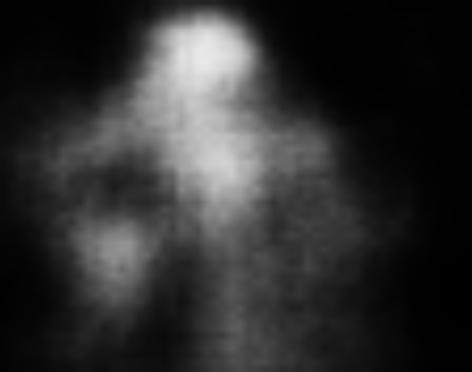}} &
		\makecell[c]{\includegraphics[width=0.061\linewidth]{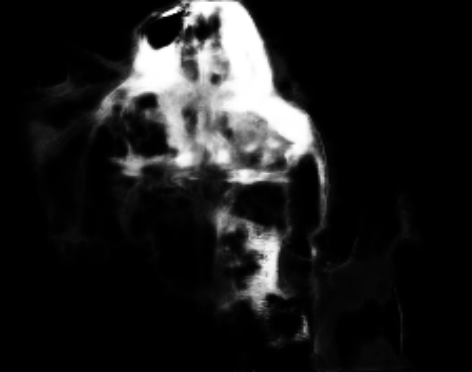}} &
		\makecell[c]{\includegraphics[width=0.061\linewidth]{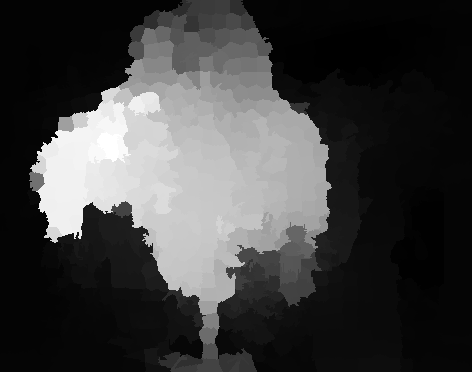}}
		\\
		II &
		\makecell[c]{\includegraphics[width=0.061\linewidth]{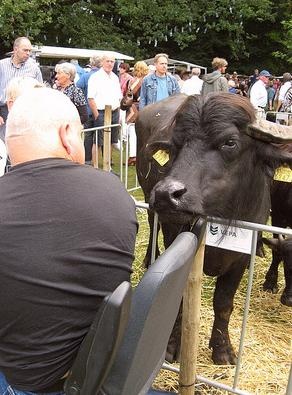}} &
		\makecell[c]{\includegraphics[width=0.061\linewidth]{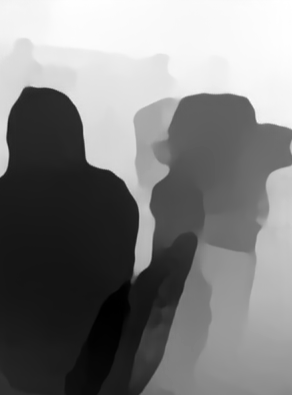}} &
		\makecell[c]{\includegraphics[width=0.061\linewidth]{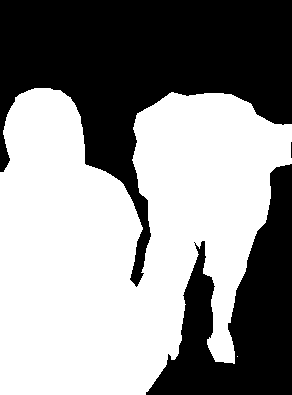}} &
		\makecell[c]{\includegraphics[width=0.061\linewidth]{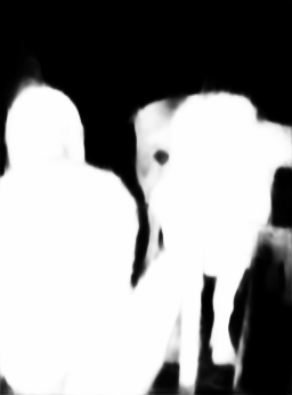}} &
		\makecell[c]{\includegraphics[width=0.061\linewidth]{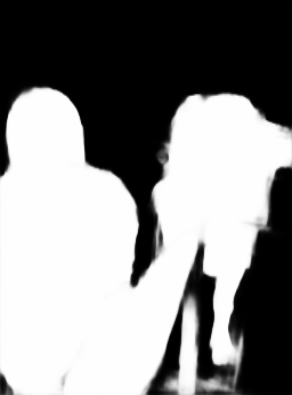}} &
		\makecell[c]{\includegraphics[width=0.061\linewidth]{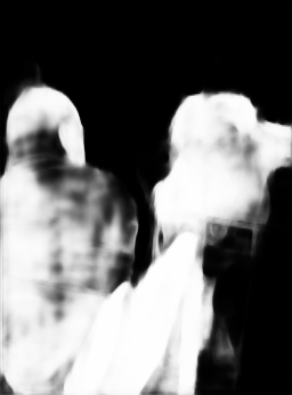}} &
		\makecell[c]{\includegraphics[width=0.061\linewidth]{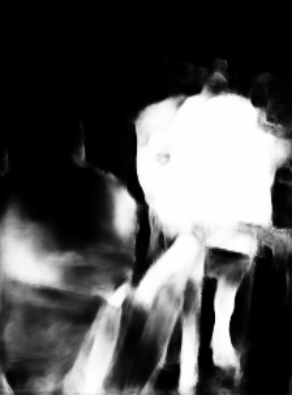}} &
		\makecell[c]{\includegraphics[width=0.061\linewidth]{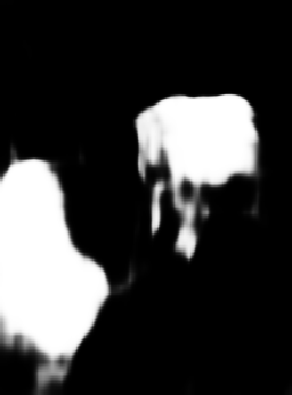}} &
		\makecell[c]{\includegraphics[width=0.061\linewidth]{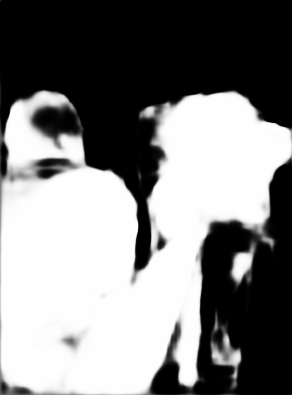}} &
		\makecell[c]{\includegraphics[width=0.061\linewidth]{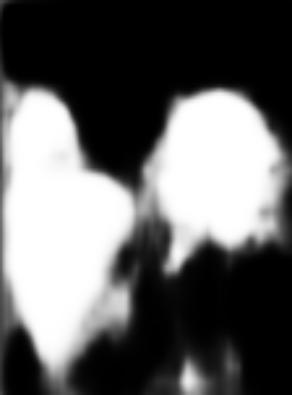}} &
		\makecell[c]{\includegraphics[width=0.061\linewidth]{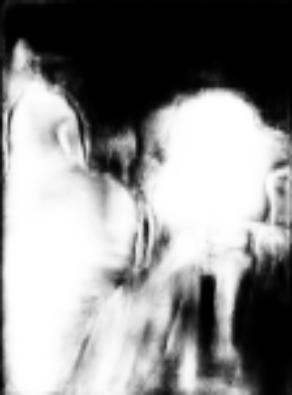}} &
		\makecell[c]{\includegraphics[width=0.061\linewidth]{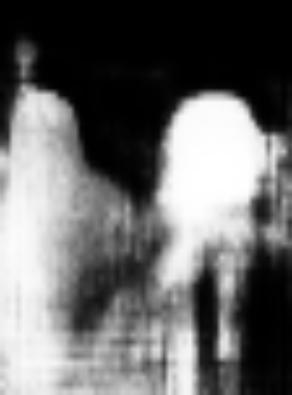}} &
		\makecell[c]{\includegraphics[width=0.061\linewidth]{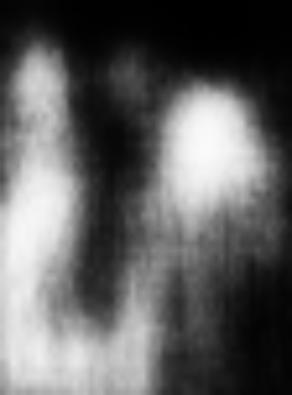}} &
		\makecell[c]{\includegraphics[width=0.061\linewidth]{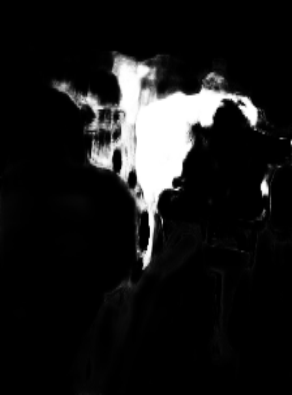}} &
		\makecell[c]{\includegraphics[width=0.061\linewidth]{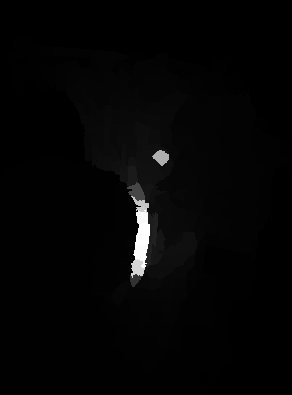}}
		\\
		III &
		\makecell[c]{\includegraphics[width=0.061\linewidth]{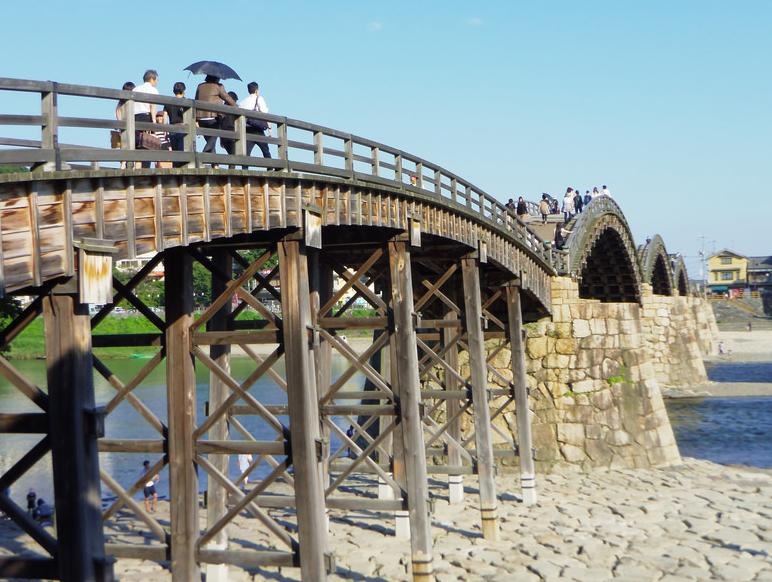}} &
		\makecell[c]{\includegraphics[width=0.061\linewidth]{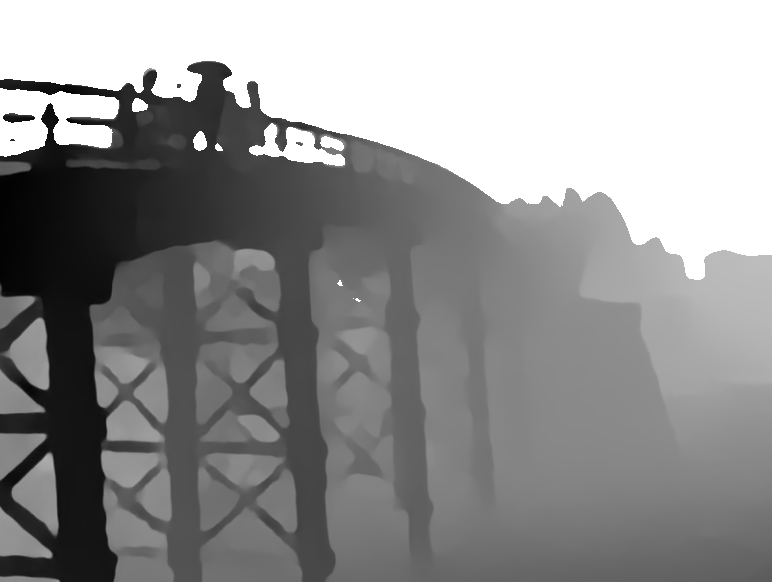}} &
		\makecell[c]{\includegraphics[width=0.061\linewidth]{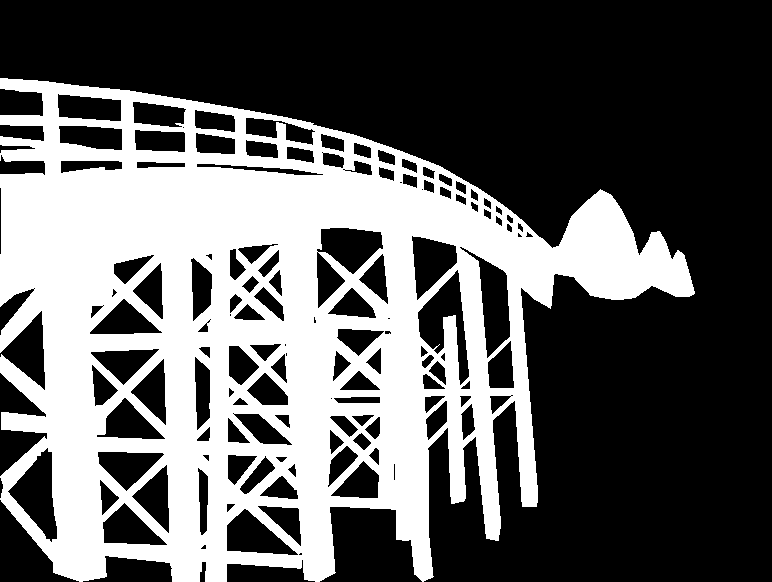}} &
		\makecell[c]{\includegraphics[width=0.061\linewidth]{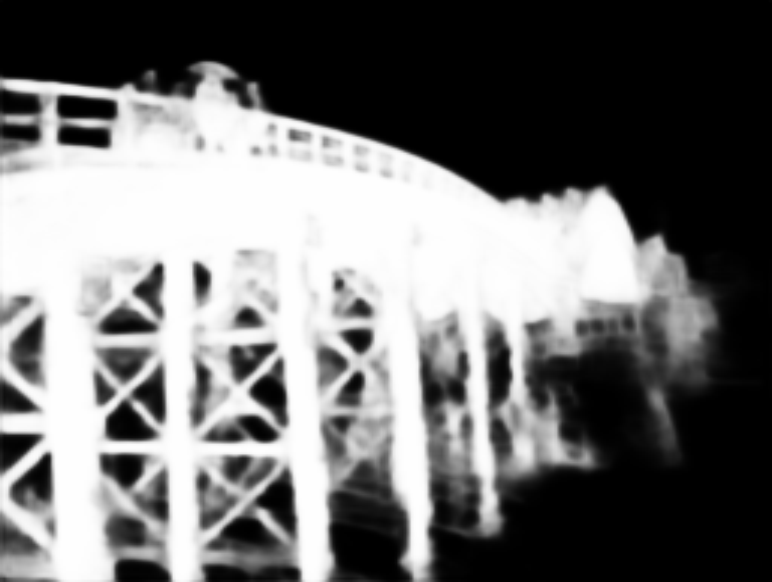}} &
		\makecell[c]{\includegraphics[width=0.061\linewidth]{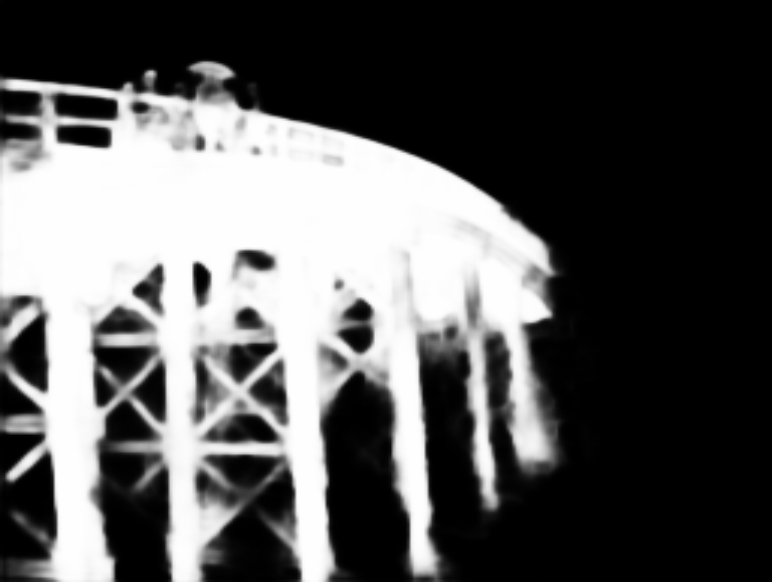}} &
		\makecell[c]{\includegraphics[width=0.061\linewidth]{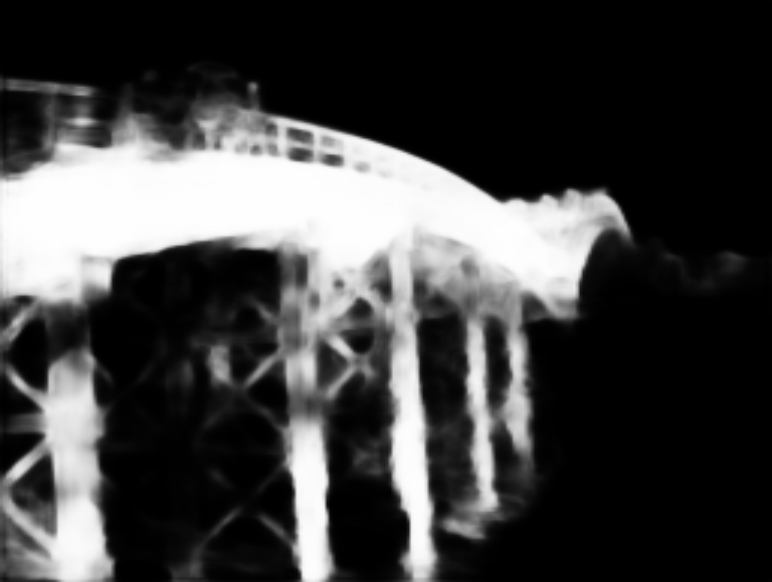}} &
		\makecell[c]{\includegraphics[width=0.061\linewidth]{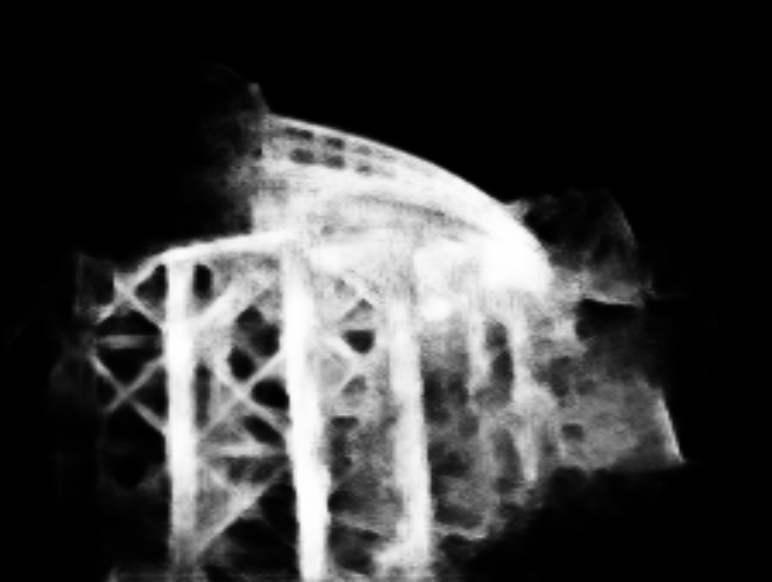}} &
		\makecell[c]{\includegraphics[width=0.061\linewidth]{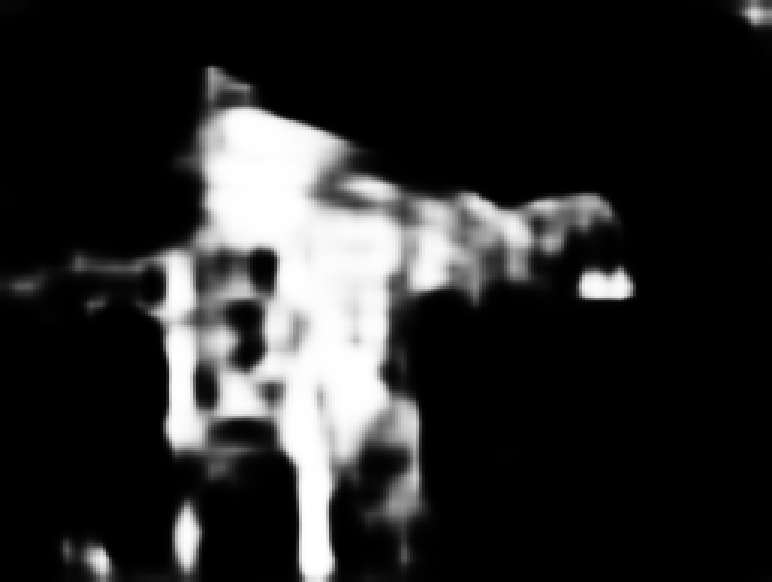}} &
		\makecell[c]{\includegraphics[width=0.061\linewidth]{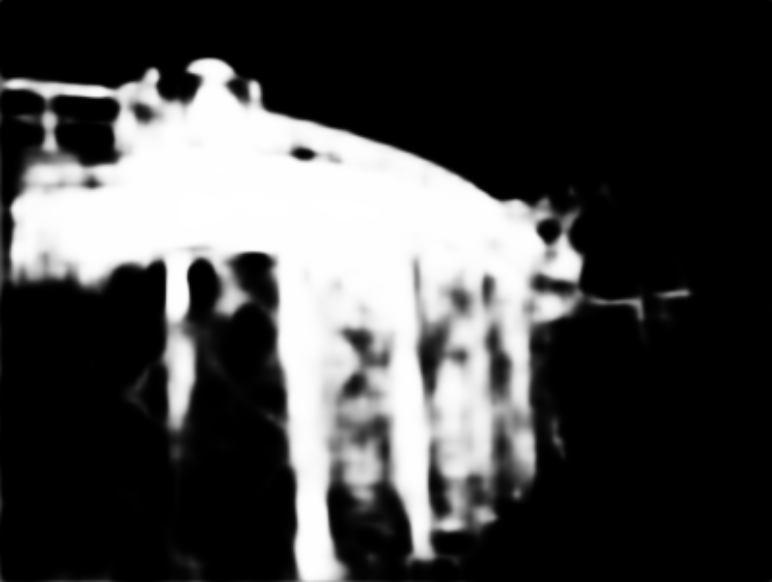}} &
		\makecell[c]{\includegraphics[width=0.061\linewidth]{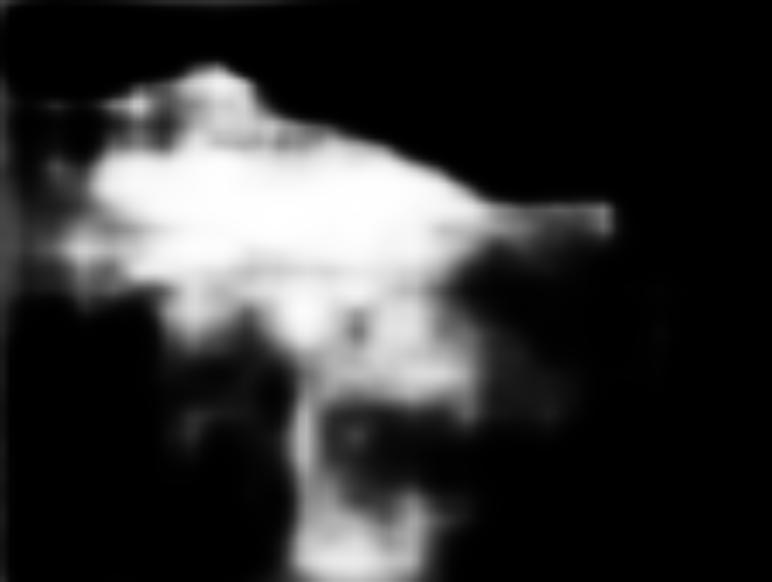}} &
		\makecell[c]{\includegraphics[width=0.061\linewidth]{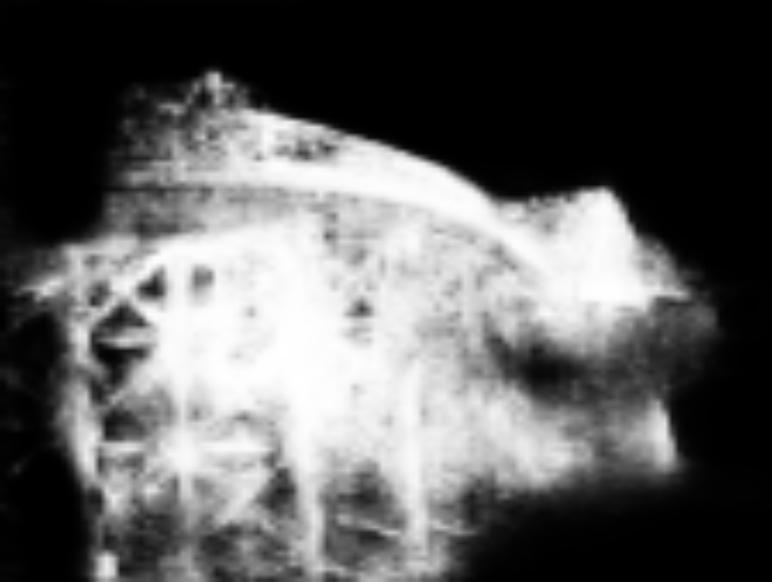}} &
		\makecell[c]{\includegraphics[width=0.061\linewidth]{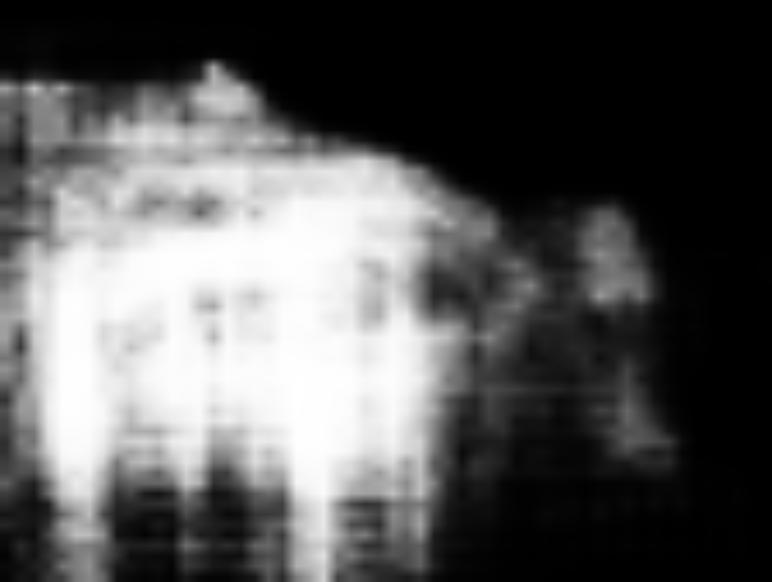}} &
		\makecell[c]{\includegraphics[width=0.061\linewidth]{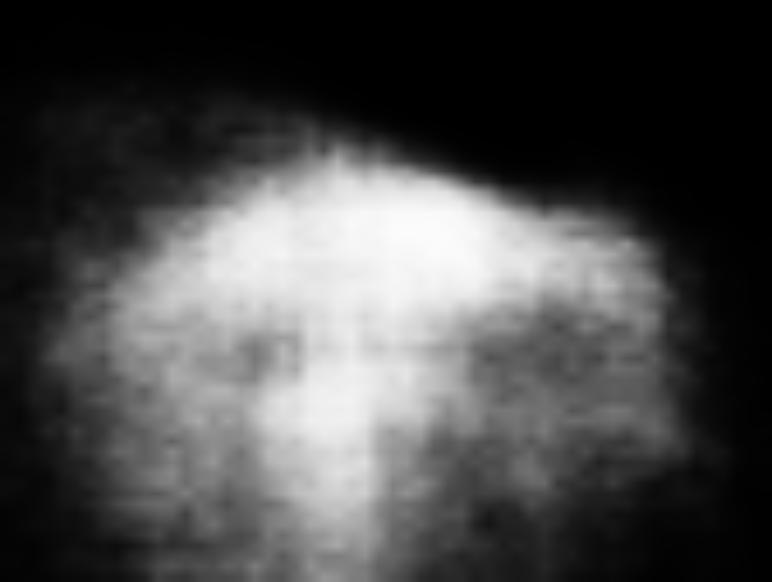}} &
		\makecell[c]{\includegraphics[width=0.061\linewidth]{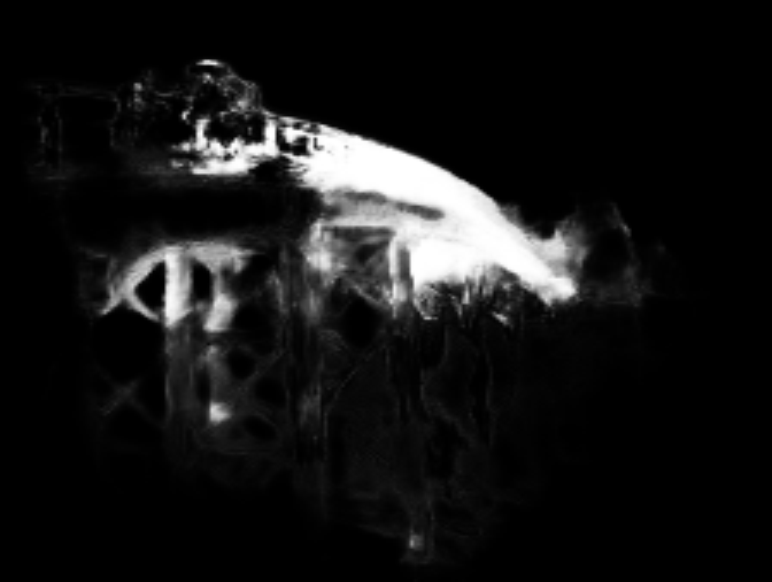}} &
		\makecell[c]{\includegraphics[width=0.061\linewidth]{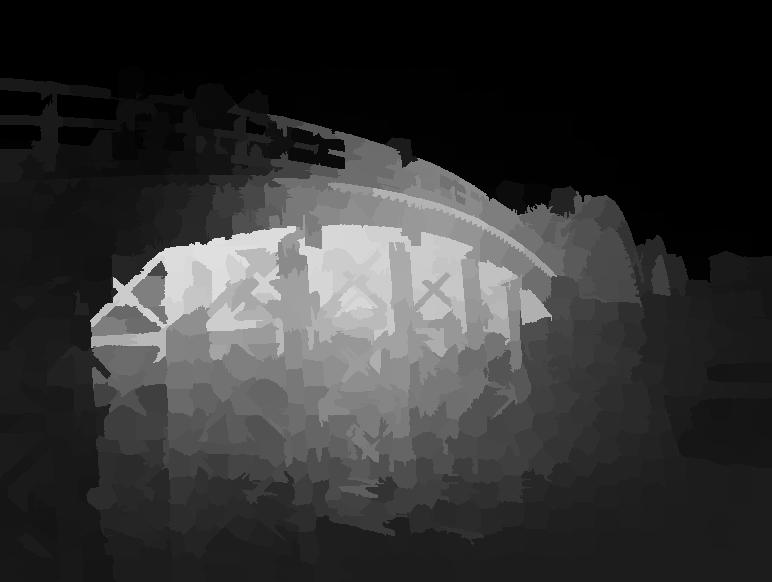}}
		\\
		IV &
		\makecell[c]{\includegraphics[width=0.061\linewidth]{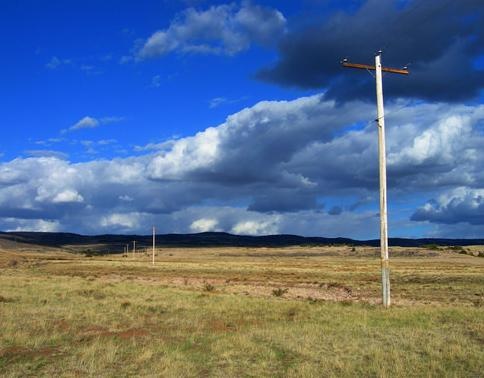}} &
		\makecell[c]{\includegraphics[width=0.061\linewidth]{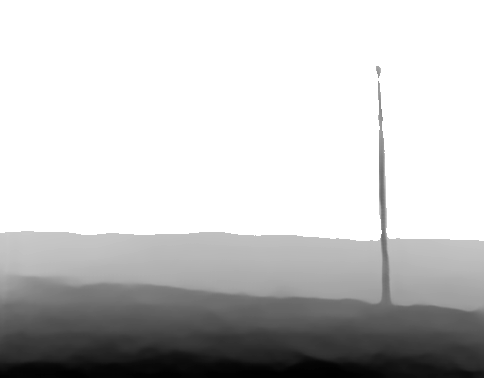}} &
		\makecell[c]{\includegraphics[width=0.061\linewidth]{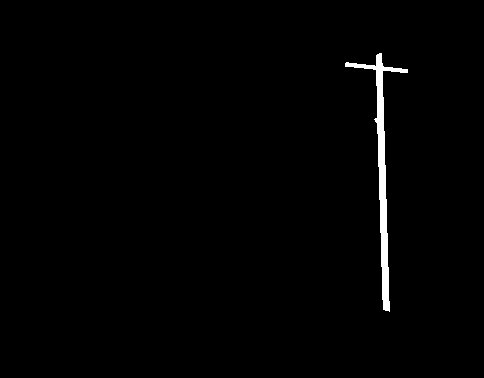}} &
		\makecell[c]{\includegraphics[width=0.061\linewidth]{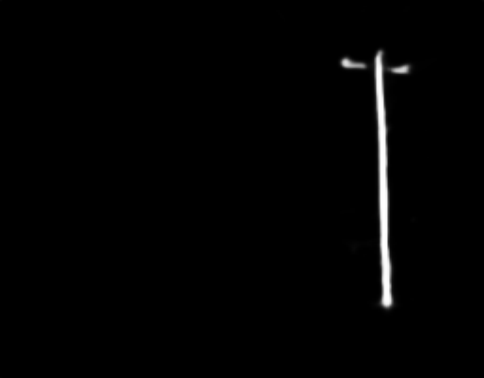}} &
		\makecell[c]{\includegraphics[width=0.061\linewidth]{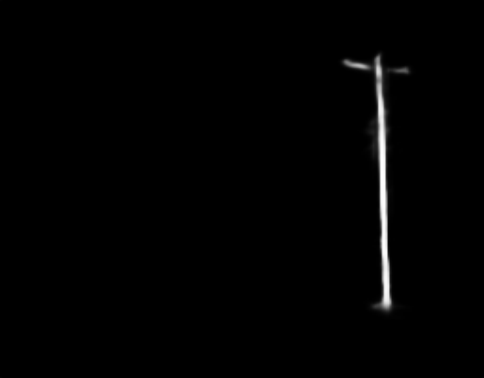}} &
		\makecell[c]{\includegraphics[width=0.061\linewidth]{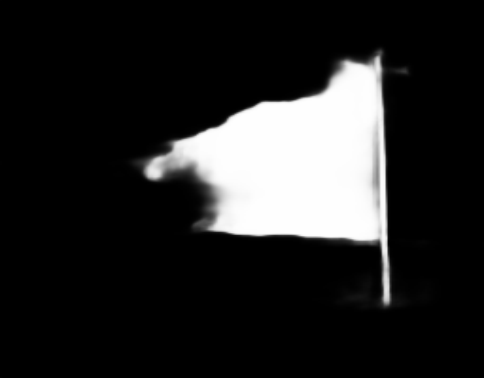}} &
		\makecell[c]{\includegraphics[width=0.061\linewidth]{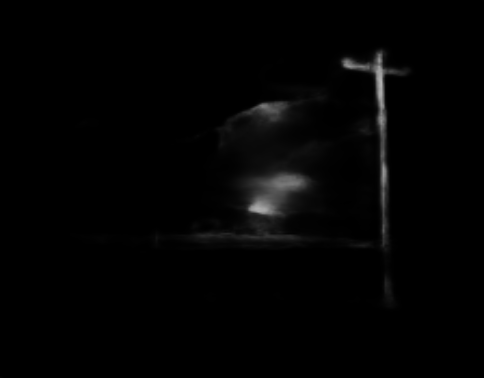}} &
		\makecell[c]{\includegraphics[width=0.061\linewidth]{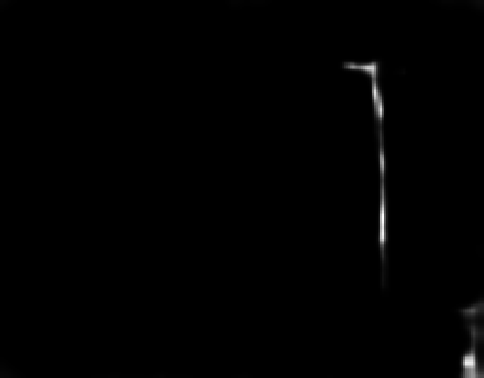}} &
		\makecell[c]{\includegraphics[width=0.061\linewidth]{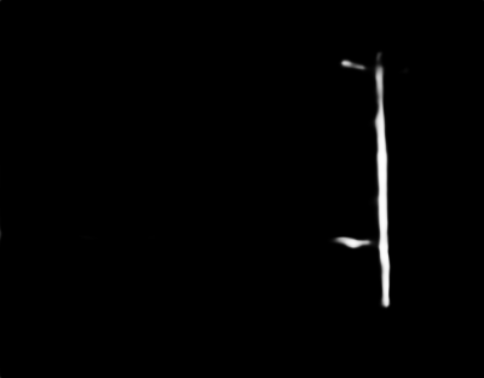}} &
		\makecell[c]{\includegraphics[width=0.061\linewidth]{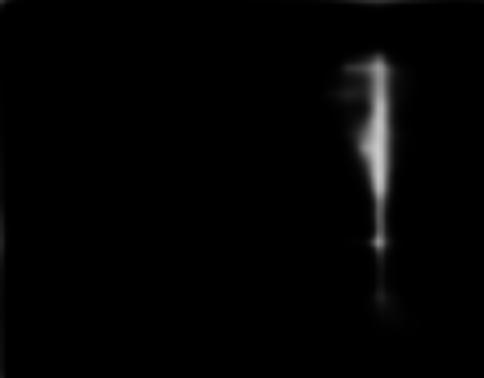}} &
		\makecell[c]{\includegraphics[width=0.061\linewidth]{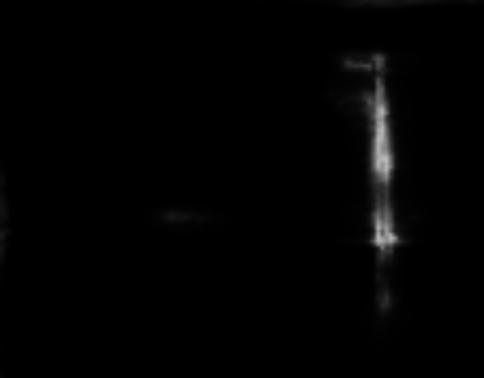}} &
		\makecell[c]{\includegraphics[width=0.061\linewidth]{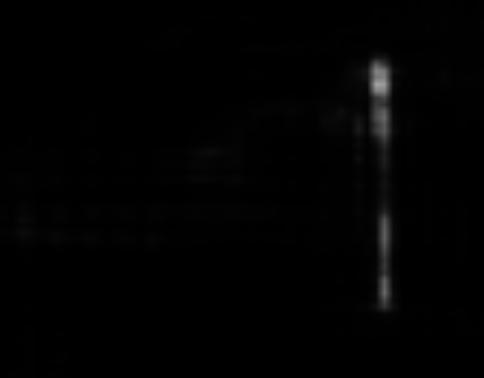}} &
		\makecell[c]{\includegraphics[width=0.061\linewidth]{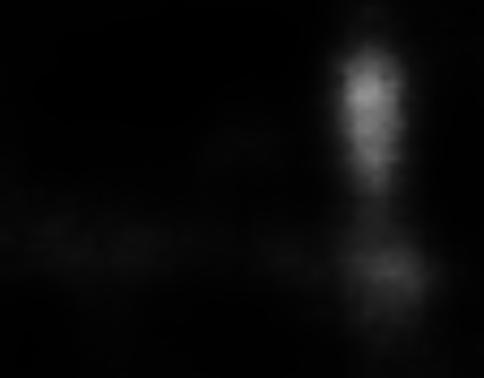}} &
		\makecell[c]{\includegraphics[width=0.061\linewidth]{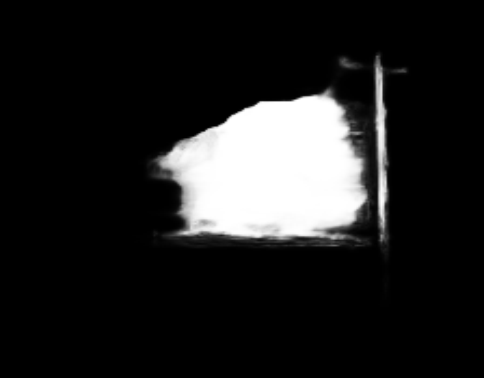}} &
		\makecell[c]{\includegraphics[width=0.061\linewidth]{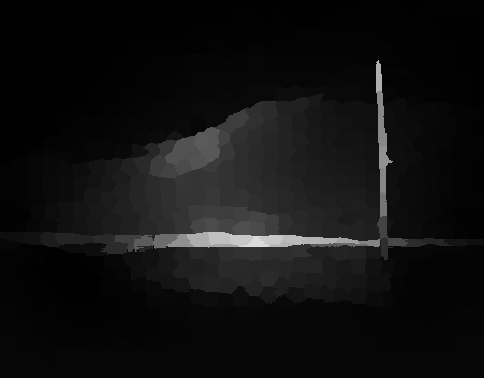}}
		\\
		V &
		\makecell[c]{\includegraphics[width=0.061\linewidth]{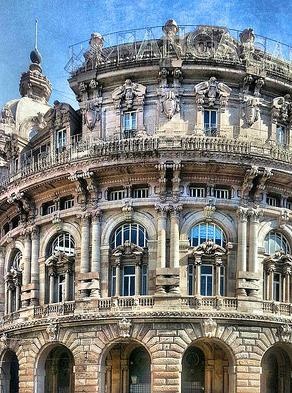}} &
		\makecell[c]{\includegraphics[width=0.061\linewidth]{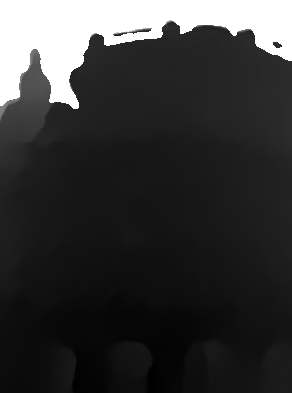}} &
		\makecell[c]{\includegraphics[width=0.061\linewidth]{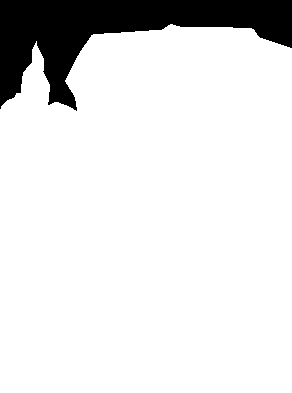}} &
		\makecell[c]{\includegraphics[width=0.061\linewidth]{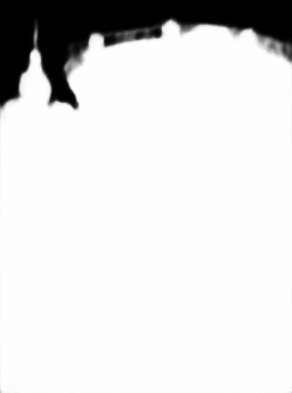}} &
		\makecell[c]{\includegraphics[width=0.061\linewidth]{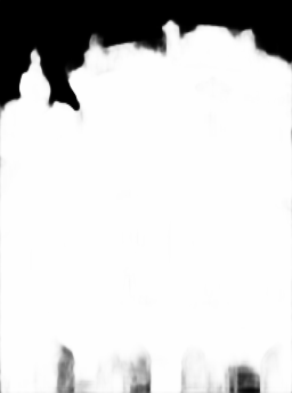}} &
		\makecell[c]{\includegraphics[width=0.061\linewidth]{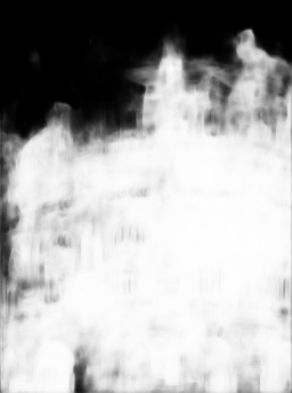}} &
		\makecell[c]{\includegraphics[width=0.061\linewidth]{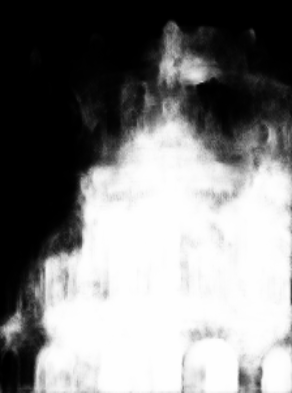}} &
		\makecell[c]{\includegraphics[width=0.061\linewidth]{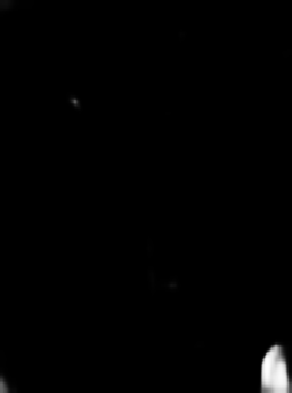}} &
		\makecell[c]{\includegraphics[width=0.061\linewidth]{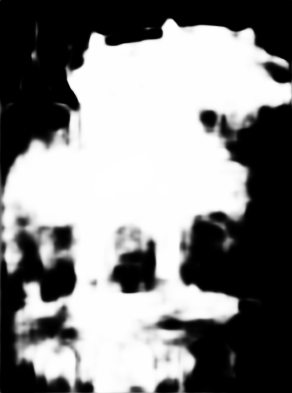}} &
		\makecell[c]{\includegraphics[width=0.061\linewidth]{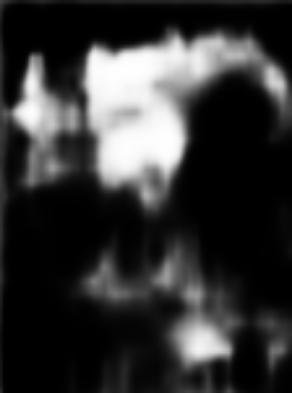}} &
		\makecell[c]{\includegraphics[width=0.061\linewidth]{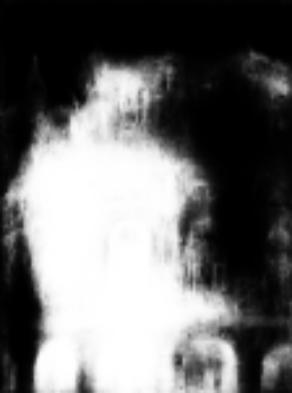}} &
		\makecell[c]{\includegraphics[width=0.061\linewidth]{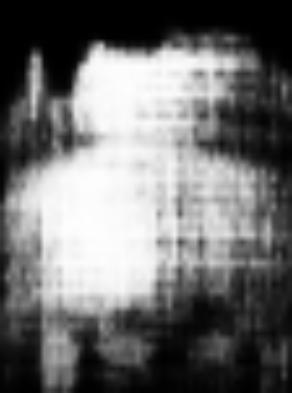}} &
		\makecell[c]{\includegraphics[width=0.061\linewidth]{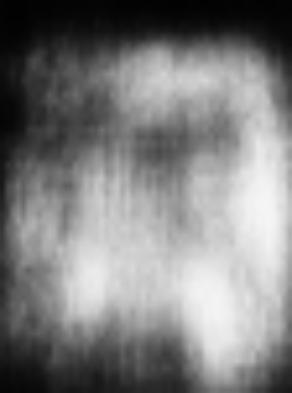}} &
		\makecell[c]{\includegraphics[width=0.061\linewidth]{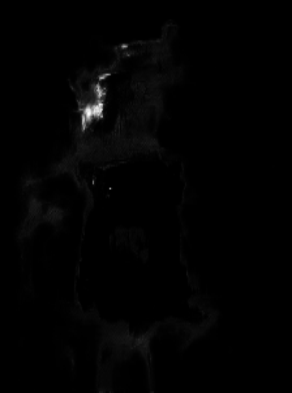}} &
		\makecell[c]{\includegraphics[width=0.061\linewidth]{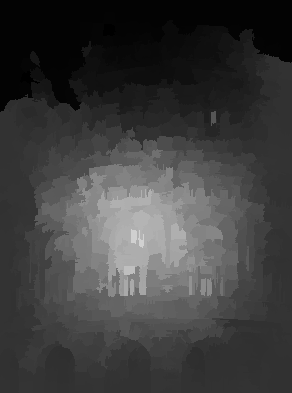}}
		\\
		VI &
		\makecell[c]{\includegraphics[width=0.061\linewidth]{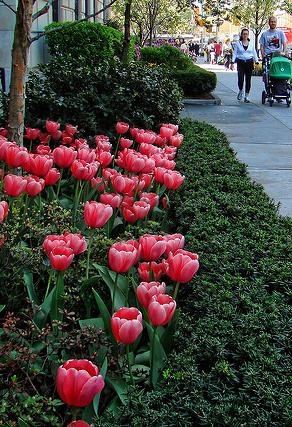}} &
		\makecell[c]{\includegraphics[width=0.061\linewidth]{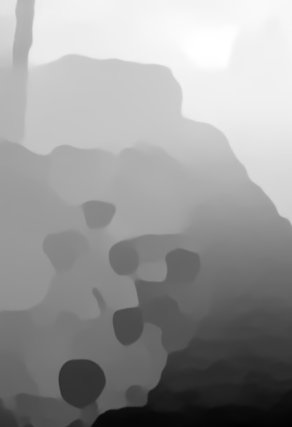}} &
		\makecell[c]{\includegraphics[width=0.061\linewidth]{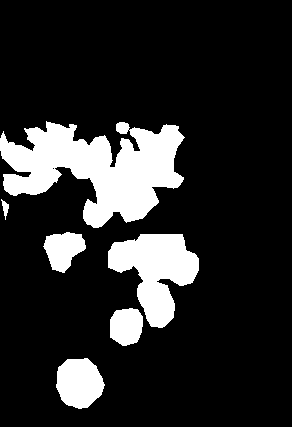}} &
		\makecell[c]{\includegraphics[width=0.061\linewidth]{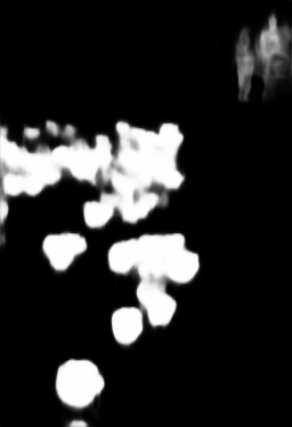}} &
		\makecell[c]{\includegraphics[width=0.061\linewidth]{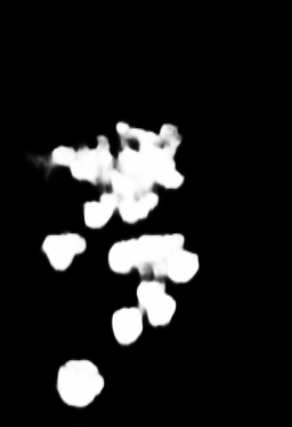}} &
		\makecell[c]{\includegraphics[width=0.061\linewidth]{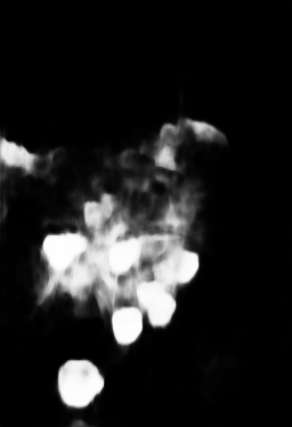}} &
		\makecell[c]{\includegraphics[width=0.061\linewidth]{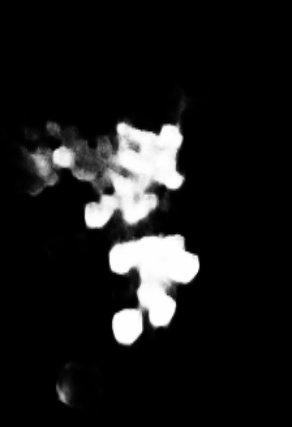}} &
		\makecell[c]{\includegraphics[width=0.061\linewidth]{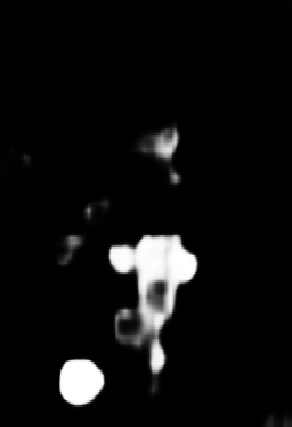}} &
		\makecell[c]{\includegraphics[width=0.061\linewidth]{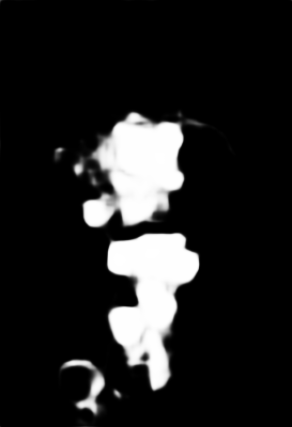}} &
		\makecell[c]{\includegraphics[width=0.061\linewidth]{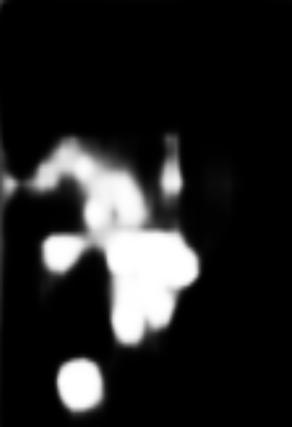}} &
		\makecell[c]{\includegraphics[width=0.061\linewidth]{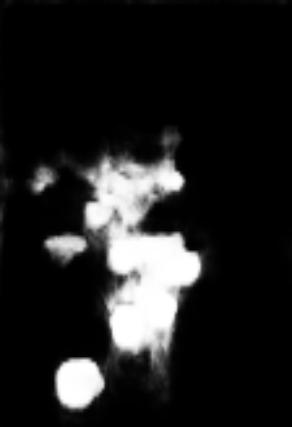}} &
		\makecell[c]{\includegraphics[width=0.061\linewidth]{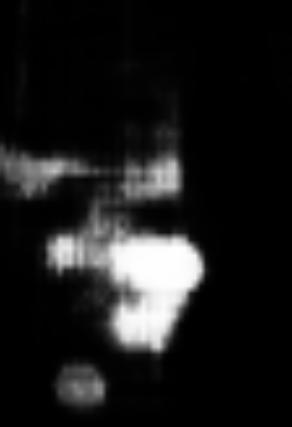}} &
		\makecell[c]{\includegraphics[width=0.061\linewidth]{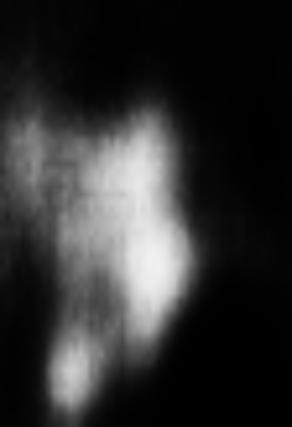}} &
		\makecell[c]{\includegraphics[width=0.061\linewidth]{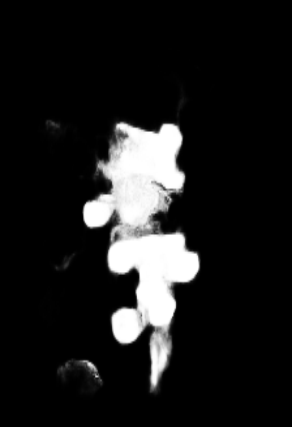}} &
		\makecell[c]{\includegraphics[width=0.061\linewidth]{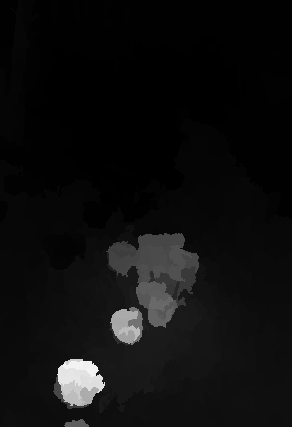}}
		\\
		VII &
		\makecell[c]{\includegraphics[width=0.061\linewidth]{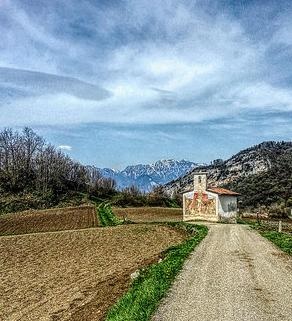}} &
		\makecell[c]{\includegraphics[width=0.061\linewidth]{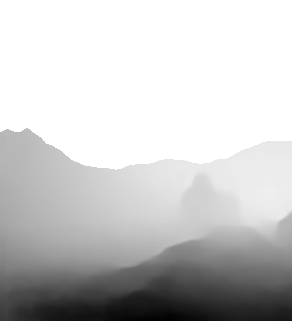}} &
		\makecell[c]{\includegraphics[width=0.061\linewidth]{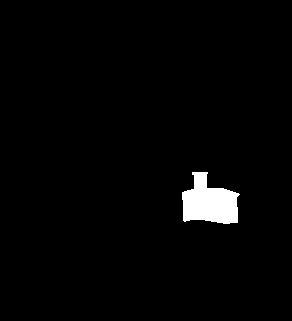}} &
		\makecell[c]{\includegraphics[width=0.061\linewidth]{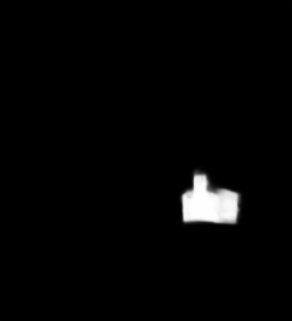}} &
		\makecell[c]{\includegraphics[width=0.061\linewidth]{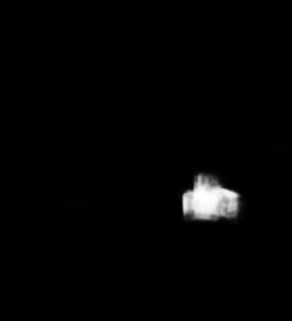}} &
		\makecell[c]{\includegraphics[width=0.061\linewidth]{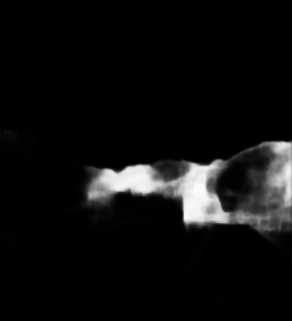}} &
		\makecell[c]{\includegraphics[width=0.061\linewidth]{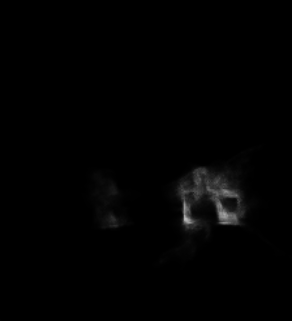}} &
		\makecell[c]{\includegraphics[width=0.061\linewidth]{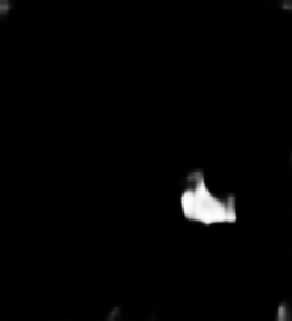}} &
		\makecell[c]{\includegraphics[width=0.061\linewidth]{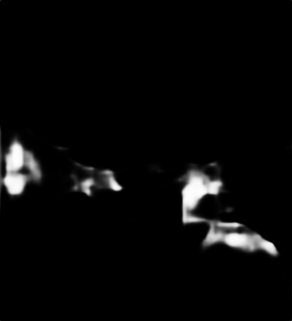}} &
		\makecell[c]{\includegraphics[width=0.061\linewidth]{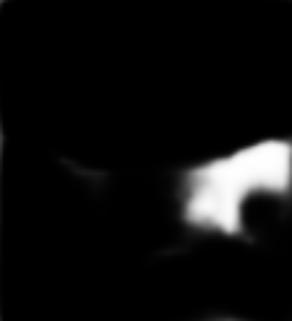}} &
		\makecell[c]{\includegraphics[width=0.061\linewidth]{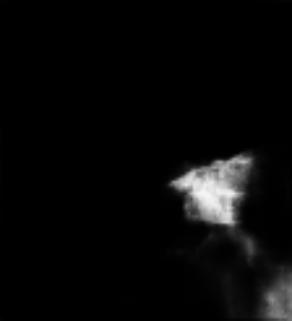}} &
		\makecell[c]{\includegraphics[width=0.061\linewidth]{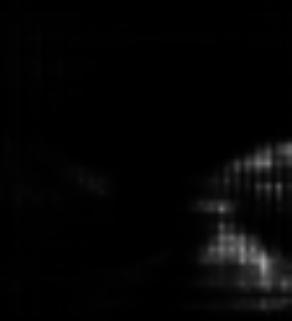}} &
		\makecell[c]{\includegraphics[width=0.061\linewidth]{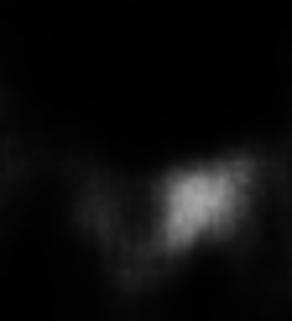}} &
		\makecell[c]{\includegraphics[width=0.061\linewidth]{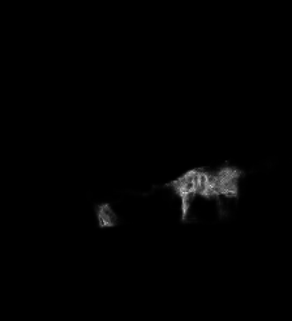}} &
		\makecell[c]{\includegraphics[width=0.061\linewidth]{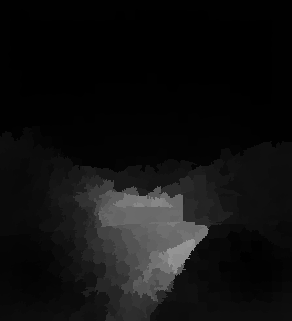}}
		\\
		VIII &
		\makecell[c]{\includegraphics[width=0.061\linewidth]{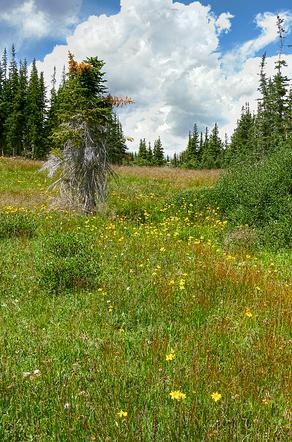}} &
		\makecell[c]{\includegraphics[width=0.061\linewidth]{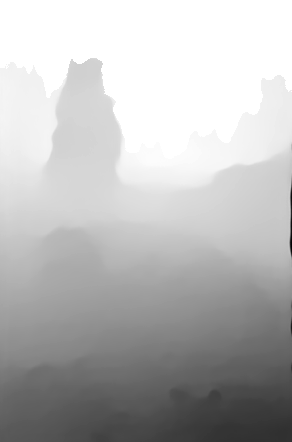}} &
		\makecell[c]{\includegraphics[width=0.061\linewidth]{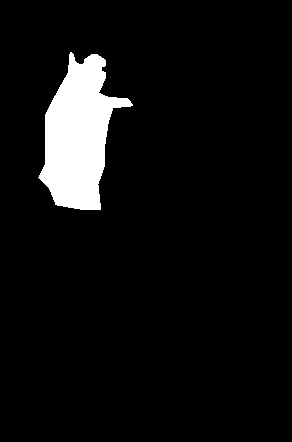}} &
		\makecell[c]{\includegraphics[width=0.061\linewidth]{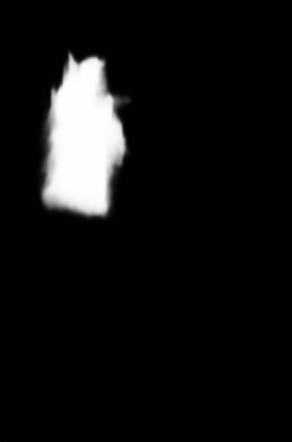}} &
		\makecell[c]{\includegraphics[width=0.061\linewidth]{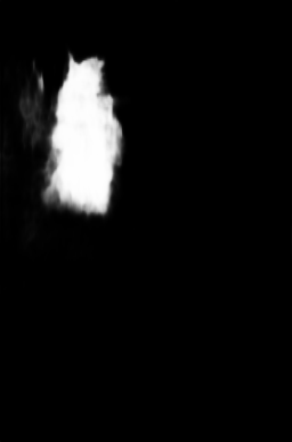}} &
		\makecell[c]{\includegraphics[width=0.061\linewidth]{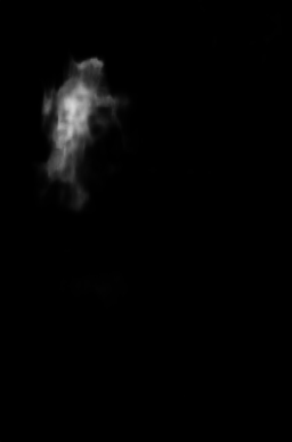}} &
		\makecell[c]{\includegraphics[width=0.061\linewidth]{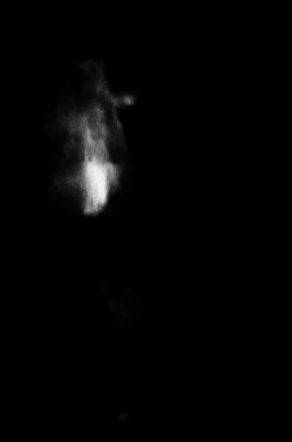}} &
		\makecell[c]{\includegraphics[width=0.061\linewidth]{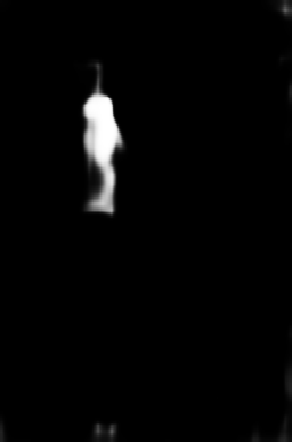}} &
		\makecell[c]{\includegraphics[width=0.061\linewidth]{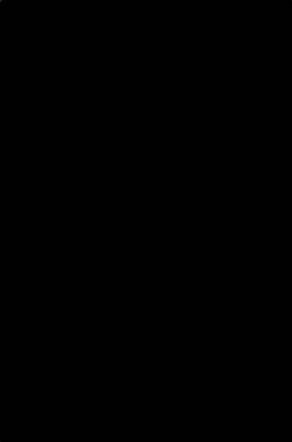}} &
		\makecell[c]{\includegraphics[width=0.061\linewidth]{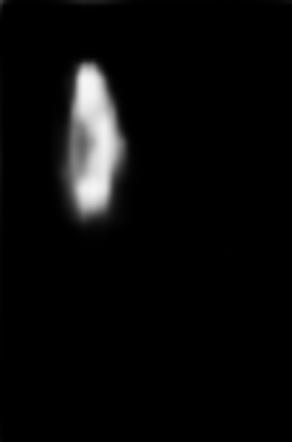}} &
		\makecell[c]{\includegraphics[width=0.061\linewidth]{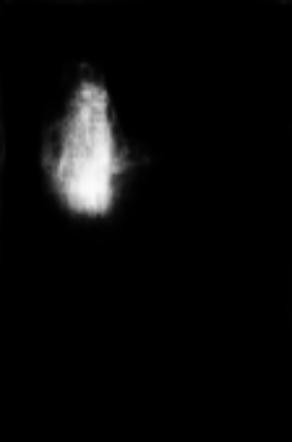}} &
		\makecell[c]{\includegraphics[width=0.061\linewidth]{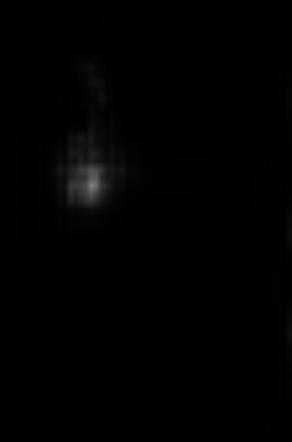}} &
		\makecell[c]{\includegraphics[width=0.061\linewidth]{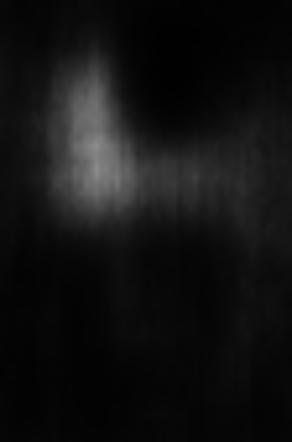}} &
		\makecell[c]{\includegraphics[width=0.061\linewidth]{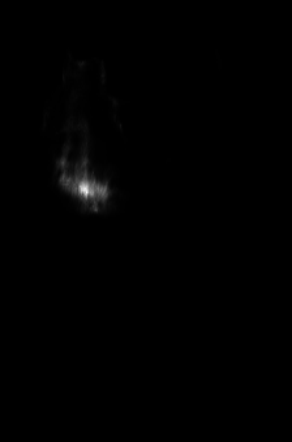}} &
		\makecell[c]{\includegraphics[width=0.061\linewidth]{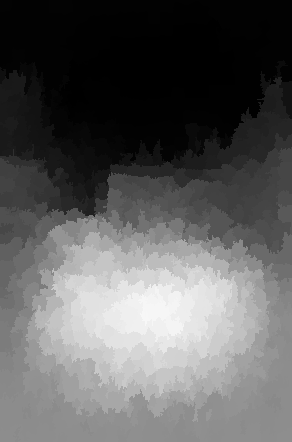}}
		\\
		&
		Image &
		Depth &
		GT&
		SMAC* &
		SMAC &
		\makecell[c]{$S^2$MA \\ \cite{liu2020s2ma}} &
		\makecell[c]{D$3$Net \\ \cite{fan2019d3net}} &
		\makecell[c]{DMRA \\ \cite{Piao2019dmra}} &
		\makecell[c]{CPFP \\ \cite{zhao2019cpfp}} &
		\makecell[c]{TANet \\ \cite{chen2019tanet}} &
		\makecell[c]{PCF \\ \cite{chen2018pcf}} &
		\makecell[c]{MMCI \\ \cite{chen2019mmci}} & 
		\makecell[c]{CTMF \\ \cite{han2017ctmf}} &
		\makecell[c]{AFNet \\ \cite{wang2019afnet}} &
		\makecell[c]{DF \\ \cite{qu2017df}}
		\\
	\end{tabular}
	\caption{
		\textbf{\textbf{Qualitative comparison against 10 state-of-the-art RGB-D SOD methods on our proposed ReDWeb-S dataset.} (GT: ground truth)}
	}
	\label{fig:visualcmp}
	\vspace{-0.3cm}
\end{figure*}

\Paragraph{Comparison with Other Models.} We first report the results of using self-attention in row VII of Table~\ref{ablationTab}. By comparing row II with it, we find that using our proposed mutual attention significantly outperforms the self-attention mechanism. We also find that directly using self-attention even downgrades the RGB-D SOD performance by comparing row I with row VII, which we believe is due to the insufficient feature discriminability. We then report the results of using our previous $S^2$MA module \cite{liu2020s2ma} in row VIII. Compared with row IV, we find that cascading MAC and NL modules outperform $S^2$MA, especially on the LFSD dataset. This result further verifies the effectiveness of our designs to not fuse self-attention with mutual attention anymore, incorporate contrast, and cascade NL right after the MAC module.

To evaluate the necessity of adopting MA modules in decoders, we report the results of using the concatenation based CMD in all five decoders in row IX. Compared with the results in row V, we observe that substituting the concatenation based fusion with MA in the first three decoders achieves large performance gains on two out of the three datasets, \ie ReDWeb-S and LFSD. This result demonstrates the superiority of adopting the MA mechanism on multi-level features.

Finally, we also compare with some naive fusion methods, such as concatenation, summation, and multiplication. We adopt them after the encoder and in every decoder module and report their results in rows X to XII. We can see that our final results in row VI outperform theirs on all the three datasets by a large margin, thus demonstrating the necessity of adopting high-order cross-modal information interactions.

\subsection{Comparison with State-of-the-Art Methods}
We compare our final SMAC RGB-D SOD model with state-of-the-art methods on all the nine datasets. Specifically, the compared methods include DF \cite{qu2017df}, AFNet \cite{wang2019afnet}, CTMF \cite{han2017ctmf}, MMCI \cite{chen2019mmci}, PCF \cite{chen2018pcf}, TANet \cite{chen2019tanet}, CPFP \cite{zhao2019cpfp}, DMRA \cite{Piao2019dmra}, D$^3$Net \cite{fan2019d3net}, and our previous $S^2$MA \cite{liu2020s2ma}. To demonstrate the benefit of our proposed ReDWeb-S dataset, we also train a SMAC model additionally using its training set and name this model as SMAC*.

The quantitative comparison results are given in Table~\ref{SOTATab}. From the comparison we observe that our new SMAC model outperforms previous methods on eight datasets. Especially, it outperforms our previous $S^2$MA SOD model by a large margin on most datasets, demonstrating the effectiveness of our extensions. Comparing SMAC* with SMAC, we can conclude that including ReDWeb-S in the training set can improve the results on most datasets, especially on SIP, which mainly contains real-world scenes. However, on RGBD135 and DUTLF-Depth, SMAC* is worse than SMAC. This is reasonable since these two datasets both mainly focus on simplex artifacts in a close range. These observations verify the benefit of our proposed ReDWeb-S dataset for real-word RGB-D SOD.

We also show the qualitative comparison of some ReDWeb-S images in Figure~\ref{fig:visualcmp}. The first three rows show three images with very complex visual scenes, such as complex foreground objects and cluttered backgrounds. We can see that such very challenging scenarios are very difficult for most previous methods while our SMAC and SMAC* models can successfully locate most parts of the salient objects. Furthermore, we also show images with small salient objects, large salient objects, multiple salient objects, faraway salient objects, and inconspicuous salient objects in rows from IV to VIII, respectively. Our models can handle all these challenging situations and outperform previous methods, demonstrating the effectiveness of our proposed SMAC model.

\subsection{Effectiveness of the ReDWeb-S Dataset}

\begin{table} [t]
	\begin{center}
		\caption{\textbf{Comparison of the average performance rank (APR) of different training settings.} Smaller APR means better performance. Due to the space limitation, we abbreviate the DUTLF-Depth dataset as DUT. Different settings have been ranked based on APR.}
		\label{tab:training_data_cmp}
		\footnotesize
		\begin{tabular}{@{}L{1.5cm}C{0.8cm}|L{2.6cm}C{0.8cm}@{}}
			\toprule
			Training Set & APR $\downarrow$ & Training Set & APR $\downarrow$
			\\ \midrule
			NJUD       & 1.69 & NJUD+ReDWeb-S & 2.72
			\\
			DUT        & 2.28 & NJUD+DUT      & 2.75
			\\
			ReDWeb-S   & 2.89 & NJUD+NLPR     & 3.36
			\\
			NLPR       & 3.08 & DUT+ReDWeb-S  & 3.42
			\\
			&      & DUT+NLPR      & 3.75
			\\
			&      & NLPR+ReDWeb-S & 4.45
			\\ \bottomrule
		\end{tabular}
		\vspace{-0.4cm}
	\end{center}{}
\end{table}

In this section, we further analyze the effectiveness of the proposed ReDWeb-S dataset by comparing the model performance of using different training sets. We first train our SMAC SOD model using one of the four training datasets, \ie NJUD, NLPR, DUTLF-Depth, and ReDWeb-S. Due to the space limitation, we do not report all the results on the nine datasets in terms of the four metrics. Instead, we follow \cite{fan2019d3net} to rank the performance of different training settings on each dataset under each evaluation metric. Finally, we compute and report the average performance rank for each setting on all the nine datasets and using the four metrics. We also evaluate the model performance of using every two of the four datasets as the training set. Table~\ref{tab:training_data_cmp} shows the results. We observe that when only using one dataset as the training set, ReDWeb-S performs not good. This is because its real-world scenes have large differences with many datasets. However, when training using two datasets, combining ReDWeb-S with NJUD achieves the best performance among all the six training settings, which demonstrates the complementary role of our proposed dataset for existing ones.

\section{Conclusion}
In this paper, we first propose a real-world RGB-D SOD benchmark dataset. Different from the numerous previous datasets, it has the most image pairs with high-quality depth maps, and the most diverse visual scenes and objects. Hence, it has both high-quality and challenging, which bring large benefits for both model training and comprehensive model evaluation. For a new and more effective way to fuse cross-modal information for RGB-D SOD, we propose a novel mutual attention model to fuse non-local attention and context features from different modalities, and also achieve high-order and trilinear modality interaction. We also incorporate the contrast mechanism and obtain a unified model. A selective attention mechanism is also presented to reweight the depth cues thus reducing the impact of low-quality depth data. By embedding the proposed SMAC model into a two-stream UNet architecture, we outperform state-of-the-art RGB-D SOD methods. We also thoroughly analyze the effectiveness of the SMAC model and the proposed dataset.


%

\ifCLASSOPTIONcaptionsoff
  \newpage
\fi



\bibliographystyle{IEEEtran}
\bibliography{mutual_attention}

\end{document}